\documentclass[11pt,letterpaper]{article}

\usepackage[letterpaper,hmargin=1.1in,top=1in,bottom=1in,footskip=0.45in]{geometry}
\usepackage[T1]{fontenc}
\usepackage{lmodern}
\usepackage{microtype}
\usepackage{amsmath,amssymb,amsthm}
\usepackage{booktabs}
\usepackage{graphicx}
\usepackage{xcolor}
\usepackage{caption}
\definecolor{linkblue}{rgb}{0.10,0.20,0.45}
\usepackage[colorlinks=true,linkcolor=linkblue,citecolor=linkblue,urlcolor=linkblue]{hyperref}
\hypersetup{pdftitle={Only Means Only: what a harmonization benchmark score averages over},pdfauthor={Imad Bulji\'c, Delaida Muminovi\'c}}
\input{glyphtounicode}
\makeatletter
\renewcommand\section{\@startsection{section}{1}{\z@}{-3.5ex \@plus -1ex \@minus -.2ex}%
  {2.0ex \@plus .2ex}{\normalfont\large\bfseries\raggedright}}
\renewcommand\subsection{\@startsection{subsection}{2}{\z@}{-3.0ex \@plus -1ex \@minus -.2ex}%
  {1.2ex \@plus .2ex}{\normalfont\normalsize\bfseries\raggedright}}
\makeatother
\title{\bf Only Means Only:\\
what a harmonization benchmark score averages over}

\author{Imad Bulji\'c\\
Polytechnic Faculty, University of Zenica\\
\href{https://orcid.org/0009-0009-2723-2315}{ORCID: 0009-0009-2723-2315}\\
\href{mailto:imad.buljic@unze.ba}{imad.buljic@unze.ba}
\and
Delaida Muminovi\'c\\
Polytechnic Faculty, University of Zenica\\
\href{https://orcid.org/0009-0002-6388-650X}{ORCID: 0009-0002-6388-650X}\\
\href{mailto:delaida.muminovic@unze.ba}{delaida.muminovic@unze.ba}}

\date{}

\begin{document}
\maketitle

\begin{abstract}
Image harmonization methods adjust the colour and lighting of an object pasted into a photograph so
that it looks native to its new scene, and they are ranked by average error on one benchmark whose test
images are real photographs with one object recoloured. One photograph can yield several such test
images, and the benchmark promises that no photograph appears in both the training and the test split.
We find that 1{,}399 photographs do, and the composites built on them are a quarter of the test set.
Duplicate checks over file names or bytes miss this, because every composite is a different file. On
that quarter the three methods we score end to end read two to five times lower error than on the rest, but most of the
gap is composition rather than the sharing: the shared composites come almost entirely from one of the
benchmark's four source collections, and inside that collection the gap falls to about one and a half
times. An official score is therefore an average over two populations, mixed in a proportion the
benchmark does not report. We release a test list of 5{,}581 composites whose construction identities do not appear in
training, with four confirmed near-duplicate pairs disclosed rather than removed, adoptable without
retraining at the price of one rescore per checkpoint. We leave open whether the sharing itself
changes what a model learns.
\end{abstract}

\section{Introduction}
\label{sec:intro}

Image harmonization has advanced largely by scoring methods on one benchmark.
iHarmony4~\cite{cong2020iharmony4} takes real photographs, alters the colour and lighting of one foreground object in each so that it looks as if it came from another scene, and releases a fixed train and test split of those altered copies. Methods are ranked by error against the unaltered photograph. The benchmark has four sub-datasets, and HCOCO, built from COCO, holds most of both released lists. A score on such a benchmark is a statement about an
evaluation population as much as about a method: it depends on which composites the test split
contains and on where those composites came from.

\begin{figure}[t]
\centering
\includegraphics[width=0.86\textwidth]{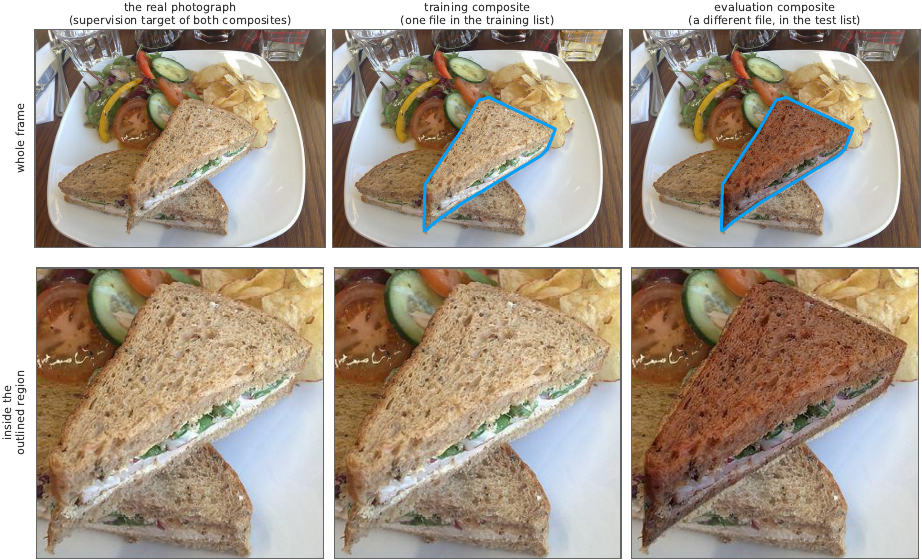}
\caption{One photograph, two composites. Both are built from the photograph on the left, the answer the model
must reproduce: the object outlined in blue has been recoloured differently in each. The two files
share no name and no bytes.}
\label{fig:phenomenon}
\end{figure}

For a composited benchmark that dependence is not incidental. One real photograph can back several
composites, and the released split decides which of them are training and which are test. Figure~\ref{fig:phenomenon}
shows the case in two composites built on one photograph, recoloured differently, one on each side
of the split. When a test composite is built on a photograph the training split also uses, the model
has already seen the pixels it is asked to reproduce, including the foreground object whose unaltered
appearance is the answer, and the reported score blends such composites with genuinely unseen ones. The standard protocol does not check for this. Its hygiene measures, such as
de-duplicating file names or file bytes, operate on composites, and two composites built from one
photograph are different files. The unit that has to be disjoint is the real photograph a composite is built from, because that
photograph is also the answer, and that unit is rarely audited before scores are compared.

We audit iHarmony4 at that construction unit. The photograph behind a composite is recovered from its
file name: the name carries the target photograph's identifier first, then the reference image used to
recolour the object, then a variant index. The official test and train lists have 1{,}399
target photographs in common, and those photographs account for 1{,}822 of the 7{,}404 composites in
the test list, 24.6 percent, each built on a photograph that also appears in training. One further composite is removed because its target photograph is pixel-identical to a training target under a different name, so
1{,}823 composites are removed in total. Removing them
leaves a retained test subset of 5{,}581 composites whose target photographs do not appear in the
training list, four confirmed near-duplicate pairs aside,
and the training list itself is left untouched, so fixed published checkpoints can be rescored
without retraining. We then run controls to determine what the sharing does and does not do to the
comparisons the benchmark is used for. Adopting the retained list is a substitution of the evaluation
list: it costs one rescore per checkpoint, and a score on it is not directly comparable with a score
on the official list.

What the sharing does to reported numbers is the harder question, and the answer is more structured
than a single effect size. Pooled over sub-datasets, the excluded composites read two to five times
lower mean squared error than the rest for the three methods scored end to end, which looks like a large advantage for a model that has seen
those photographs. Inside HCOCO, where most excluded composites sit, the same ratios fall to 1.47 to
1.61, and the DoveNet-Harmonizer gap narrows there instead of widening (post-hoc). The ratios are of whole-image mean squared error (MSE), and the
foreground-region version of the same metric, computed over the pasted object's region and written
fMSE here, is the one harmonization papers report.
A foreground-ratio-stratified control in MSE is negative for Harmonizer and for our own editor and
is not detected for DoveNet, and inside HCOCO, where the mixture cannot differ, none of the three is
detected at floors of $5.9$ to $15.7$ MSE. One published checkpoint,
DiffHarmony~\cite{zhou2024diffharmony}, moves the other way on MSE and scores worse on the excluded
composites. The pooled difference is therefore largely arithmetic over two differently composed
populations, 99 versus 44 percent HCOCO, rather than a measurement of what the sharing does to a
method.

What the sharing does to what a model learns is not measured here. A
pre-registered audit of four published checkpoints ran below its registered five-checkpoint floor,
with the retained-versus-excluded interaction significant for all six method pairs after Holm
correction on the pooled lists.
The pre-registered exposure retrain departed from its registered execution, and both of its intervals include zero (Section~\ref{sec:e1observational}).  The evaluation itself can be corrected today: the retained list is ready to adopt. 

\textbf{Contributions.} \textbf{(1)} A construction-unit audit of the released iHarmony4 lists, with
deterministic test-side exclusions and a perceptual residue check whose blind spots we state
(Sections~\ref{sec:leak} and~\ref{sec:perceptual}). \textbf{(2)} A construction-identity-disjoint
retained test subset and its excluded counterpart, kept beside the official list so that fixed
published checkpoints are rescored without retraining, together with the five evaluation-hygiene rules
the audit motivates (Sections~\ref{sec:release} and~\ref{sec:hygiene}). \textbf{(3)} A controlled
score-sensitivity analysis, reported as observational: fixed-checkpoint population contrasts,
foreground-ratio-stratified controls, a below-floor four-checkpoint audit, and an exposure retrain
whose arms departed from the registered execution and which therefore establishes neither a difference
nor practical equivalence, each with the composition and execution limits stated beside the result
(Section~\ref{sec:changes}). Tables and derivations
are in the supplement.
Section~\ref{sec:repro} states what regenerates.

\section{Evaluation populations}
\label{sec:why}

Removing the composites built on a shared photograph from a test list changes two things at once, and a reported
score alone cannot say which of them produced the change. The first is the mixture. The composites built on a shared
photograph are overwhelmingly HCOCO, so the retained list is not the same
benchmark with fewer errors: it is a differently weighted benchmark, and a method that is stronger on
HCOCO will move for that reason alone. The second is the sharing itself. On the retained list no composite is
built on a photograph the model has seen, so a method that has effectively memorized the photograph it
must reproduce loses that advantage. Both would lower the score of such a
method, and they are not separable from the score. The mixture can be held fixed inside one
sub-dataset, but the evidence against the sharing is the absence of a detected
difference, and an absence of detection is not an absence of difference.

A third possibility sits between them, and it is easy to miss because it appears as a difference of
convention rather than of population. Two evaluations of one checkpoint on one list can report
different numbers when they disagree about the metric definition, the peak used in PSNR, the working
resolution, the region of integration, the compositing rule or the batch-normalization mode. None of
those is a property of the data, and each can be as large as the effect we set out to measure.

We separate the three by construction rather than by argument. For the mixture we compare inside a
single sub-dataset, where the mixture cannot differ, and we stratify by the foreground ratio at which
the metric is evaluated, the fraction of the frame covered by the recoloured object, (Sections~\ref{sec:inflation} to~\ref{sec:probe}). The convention question is not a rival explanation: it is a property of our
measurement, and it is answered by a reproduction check that opens the results. For the
sharing itself no fixed-checkpoint analysis can substitute for an intervention, because every checkpoint
we score was trained with the sharing present.
We ran the intervention and report it with its
execution departures (Section~\ref{sec:retrain}).

\section{Shared photographs in the official lists}
\label{sec:leak}

One real photograph can back several composites. The original dataset description
promises that a target image is not shared across splits: ``We ensure that the same target image
does not appear in the training set and test set simultaneously, to avoid that the trained model
simply memorize the target image''~\cite{cong2020iharmony4}.  The released lists contain composites that stand on the same real photograph across the two
splits: the shared element is the construction identifier carried in the filename, so what
crosses the split is the real photograph the composite is built from, which is also the answer the
model must reproduce, not the composite file itself. A duplicate check over file names or file bytes does not see it, because the composites
are different files. Overlap between evaluation and training data is a known problem in
classification, where duplicates and near-duplicates inflate measured accuracy
~\cite{barz2020purging,recht2019generalize}, and it is one of the leakage classes that destabilise
published results~\cite{kapoor2023leakage}. Supplement~S10 gives a worked example with its positions in the released lists. The overlap is a property of the two lists, and the construction explains how it arises: a composite
is one target photograph paired with one reference image, the 42{,}828 HCOCO composites are built on
28{,}938 target photographs, so a photograph used as a target more than once carries more than one
composite, and the released lists place some of those composites on each side of the boundary. The
promise quoted above names the target image, and this is the case it rules out. The counts below come from the released training and test lists as the dataset repository
distributes them.
Supplement~S10 records the two list files and their content hashes. Our matcher
derives the target-photograph identity of every composite from these lists and refuses to proceed if the
layout it is handed still shares identities across splits. Section~\ref{sec:repro} states which counts this reproduces and which it does not.

We call a composite shared when the photograph it is built from appears in both lists. 1{,}399
target photographs are shared, and the composites built from them account for 1{,}822 of the 7{,}404
composites in the official test list, so 24.6 percent of the test set is built on a photograph that
also appears in training. Removing them leaves a retained test subset of 5{,}581
composites with no target photograph in common with the training list, and the training list is left
untouched, so published checkpoints stay eligible for rescoring without retraining. One further
composite is removed only by an exact decoded-pixel duplicate check, which finds exact decoded-pixel copies only. The per-sub-dataset decomposition, the validation split carved from the training
list under the same identity rule, and the staged counts are in Supplement Table~S12.
The removal
is deterministic list arithmetic, and the excluded composites are the set difference between the
official and retained test lists.

\paragraph{Other composited benchmarks.}
A post-hoc construction scan of public split lists (name level, CPU only, not a pre-registered
audit) did not find the iHarmony4 pattern in the two other benchmarks it could scan.
ccHarmony~\cite{niu2023ccharmony} (3{,}080 train and 1{,}180 test composites over 250 and 100
target photographs) shares 0 identities between train and test and 0 with the iHarmony4 official
train identities, and no real or mask file crosses its split. DESOBA~\cite{hong2022desoba} (839
train and 160 test source images) shares 1 scene stem, 1 of 160 test images (0.6 percent), read
as near-duplicate scene content (shifted viewpoint or crop, grayscale correlation 0.35, one pair
inspected by eye).  iHarmony8K, DESOBAv2, SRD, USR and WSID were blocked (no source found, or
archive too large to fetch).
ISTD+ and HYouTube were not scanned (Supplement Table~S11).
The scan does not test near-duplicates under different names, pretraining overlap or scores.

\subsection{Limits of identity matching}
\label{sec:perceptual}

The matcher keys on identity strings, so two frames of one scene, two crops of one photograph,
or a re-encoded duplicate under a different filename would not be matched. A 64-bit dHash over
every unique target photograph (pool sizes and the full radius table in Supplement
Table~S7), test queried against train at Hamming radii $\{0, 2, 4, 6, 8\}$ with every Hamming-0
pair confirmed by a $64 \times 64$ grayscale correlation, returns at Hamming 0 four cross-split
pairs beyond the registered exclusions (1 HCOCO, 1 HAdobe5k, 2 HFlickr): three are identical
content ($r \geq 0.9999$, recompression under a different filename) and one is same-scene
($r = 0.9509$). These are 4 of 2{,}917 target photographs in the retained test list, a confirmed floor of $0.14$
percent and not the size of the residue. They remain in the retained list. Wider radii return
candidates (5 at $\leq 2$ bits, 8 at $\leq 4$, 19 at $\leq 6$, 78 at $\leq 8$) that were not
adjudicated. The retained list is therefore disjoint by path identity only and is not a clean
split.  A whole-matrix screen extends
that check beyond the hash radii: every one of the 7{,}404 test composites is compared with
every one of the 65{,}742 training and validation composites at $32 \times 32$ grayscale
correlation,
and cross-split pairs above $0.99$ whose identity strings differ are reported. It returns six
candidates. One is byte-identical, and its composite is one of the composites removed by the decoded-pixel
check.
The
other five sit in the retained list at correlation $0.99997$ or above, so the retained list
carries them (Supplement Section~S4). 

\section{Effect on reported scores}
\label{sec:changes}

Before reading any difference as population, we fixed the conventions, because two evaluations of one
checkpoint on one list can disagree without any population changing. Our scoring reproduces the numbers
its authors published: to within $0.13$ MSE for DoveNet once its upstream batch-normalization setting
is used, and to within $1.2$ percent fMSE for Harmonizer under its authors' own evaluation code, and every
table in this paper uses one protocol for every method (Supplement Sections~S18 and S17).

How large is the difference
between the two populations in the reported metric? How much of that difference is composition
rather than the sharing? Does it reproduce across methods and inside a single sub-dataset, and does
it survive a control that matches the foreground ratio at which the metric is evaluated? Does an
audit or an intervention establish it? And does an external benchmark confirm the retained list's
advantage? 

\subsection{Pooled error on the shared subset}
\label{sec:inflation}

\begin{table}[htbp]
\centering
\small
\begin{tabular}{lrrrr}
\toprule
Method (MSE $\downarrow$) & ret. & excl. & offic. & $\Delta$ \\
 & 5{,}581 & 1{,}823 & 7{,}404 & excl.$-$ret. \\
\midrule
DoveNet & $140.18$ & $69.40$ & $122.75$ & $-70.78$ \\
Harmonizer & $133.42$ & $26.43$ & $107.08$ & $-106.99$ \\
Editor (ours) & $149.45$ & $48.82$ & $124.68^{\dagger}$ & $-100.63$ \\
\bottomrule
\end{tabular}
\caption{Whole-image MSE on the retained, the excluded and the official list; the shared quarter reads $2$ to
$5$ times lower for every method. $^{\dagger}$Weighted mix of the two lists rather than a separate
scoring (Supplement Table~S19 in Section~S14).}
\label{tab:inflation}
\end{table}

\begin{figure}[t]
\centering
\includegraphics[width=\textwidth]{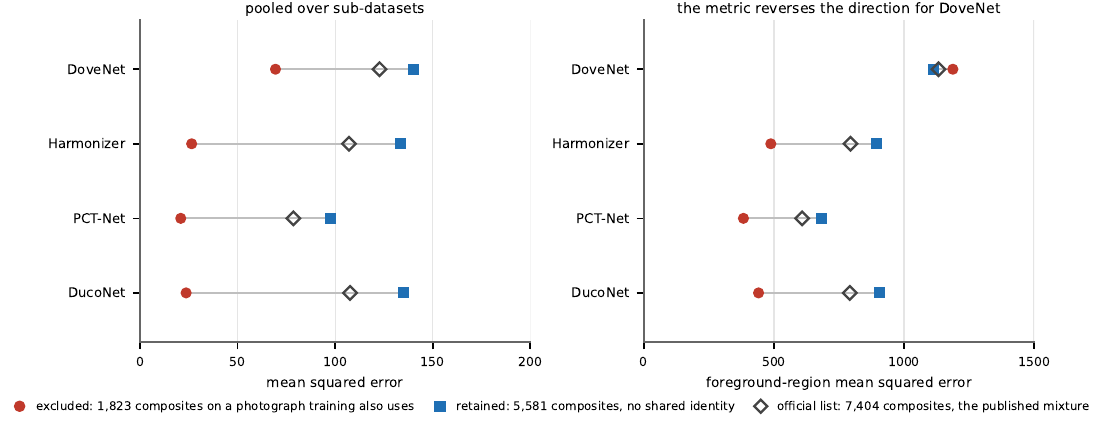}
\caption{An official score is a mixture of two populations. Each checkpoint is plotted on the excluded
composites (circle), the retained composites (square) and the official list (diamond), in whole-image
MSE at the left and foreground-region MSE at the right.}
\label{fig:population}
\end{figure}

Every method's official-list number mixes roughly 75 percent retained and 25 percent shared
composites. Pooled, the shared quarter scores $2.02$, $5.05$ and $3.06$ times lower in MSE
(DoveNet, Harmonizer, ours).
The excluded list is 99 percent HCOCO (1{,}807 of 1{,}823), the
retained list 44 percent (2{,}476 of 5{,}581). Inside HCOCO the factors are $1.47$, $1.61$ and
$1.58$ with intervals above 1, not foreground-matched (post-hoc, Supplement Table~S19).  Figure~\ref{fig:population} draws that arithmetic for the four checkpoints whose
scores are recorded on all three lists: the official-list number sits between the two populations it
averages, and on the foreground-region metric the direction turns over for DoveNet.

The direction is not shared by every checkpoint we scored. Three further published checkpoints,
scored for the audit of Section~\ref{sec:e1observational}, give retained-to-excluded MSE ratios
of $4.68$ for PCT-Net~\cite{guerreiro2023pctnet} and $5.73$ for DucoNet~\cite{tan2023duconet}
($1.56$ for both inside HCOCO), and $0.83$ for DiffHarmony ($0.90$ inside HCOCO): DiffHarmony
scores worse on the shared composites in both populations (Supplement Table~S19).
PCT-Net and DucoNet were subsequently scored on the official list as well, so all three
populations are now measured for them: their official-list MSE is $78.56$ and $107.60$, between
their retained values ($97.41$ and $135.04$) and their excluded values ($20.83$ and $23.58$),
which is the mixture arithmetic this section describes, on two further checkpoints (Supplement
Table~S5).
DiffHarmony is outside the frozen four-checkpoint audit family and is reported descriptively.
It
is built on a pretrained Stable Diffusion inpainting model whose pretraining data a list-based
audit cannot examine.

The fMSE column moves differently and we state it: DoveNet is $+73.0$ fMSE \emph{worse} on the
excluded composites, Harmonizer $-406.5$ better, ours $-233.0$ better. That is consistent with
the compositing-metric asymmetry of Section~\ref{sec:asymmetry}: the excluded composites carry
smaller foregrounds on average, and fMSE normalizes by exactly the region that is smaller.

\paragraph{Source of the pooled difference.}
\label{sec:gaps}

On the retained list the DoveNet-minus-Harmonizer MSE gap is $6.76$.
On the official list the
same two checkpoints, scored by the same driver, read $15.67$, $2.3$ times larger. This pooled
ratio is not a measure of exposure: its interval is uninformative ($[-13.42, +15.16]$).
Inside
HCOCO the official-to-retained ratio of that gap is $0.88$ $[0.84, 0.94]$ ($51.77$ against $58.73$), and the excluded-list gap is $42.25$, lower than both, so the shared composites narrow it. Removing $1{,}807$ random
HCOCO composites (plus the 16 other excluded ones) gives pooled ratios with median $4.25$ ($3.05$
to $7.57$), above the observed $2.3$ (post-hoc, a reference rather than a matched placebo, Supplement
Table~S20 in Section~S14). Between these two baselines Harmonizer leads on all three lists and no ranking flip
occurs.

\paragraph{Foreground-ratio stratification and its floor.}
\label{sec:dose}

For each test composite, the dose $k$ counts the training composites built on the same target
photograph, so $k = 0$ on the retained list by construction and $k \geq 1$ on the
excluded one (the dose distribution over the 7{,}404 composites is in Supplement
Section~S1).\footnote{The $k = 0$ count is 5{,}582 rather than 5{,}581 because the one
decoded-pixel-duplicate exclusion (its target is pixel-identical to a train target under a
\emph{different} filename) never occurs in the train identity list and so has $k = 0$.
The
remaining $1{,}822$ exclusions are path-identity matches with $k \geq 1$.} Composites are binned
on foreground ratio (eight pooled quantile bins), and the contrast is the within-bin $k \geq 1$
versus $k = 0$ step. The estimator uses list membership only: doses above 1 are not used, and
sub-dataset is not matched (Section~\ref{sec:three}).

For Harmonizer the unmatched pooled step ($k \geq 1$ minus $k = 0$) is $-106.98$ MSE, and the
within-bin step is negative in every bin (8/8). This is not the $-106.99$ of
Table~\ref{tab:inflation}, which splits the composites 1{,}823 / 5{,}581 instead of 1{,}822 /
5{,}582. For DoveNet the pooled step is $-70.90$, but within bins every step reverses sign (8/8
positive), a direct demonstration that the pooled contrast was population composition (a
negative step means shared composites score better, per-bin ranges in Supplement
Table~S1). The same reading applies to the pooled matched steps (Section~\ref{sec:three}).
Within-dose slopes are unstable in sign for both methods at these sample sizes and no trend is
claimed from them.

\subsection{Pooled and within-HCOCO steps}
\label{sec:three}

All three methods go through the identical statistic: dose $k$ as above, the bin-averaged
$k \geq 1$ minus $k = 0$ step within eight foreground-ratio bins, and a cluster bootstrap over
target photographs, with the bin-averaged step, its interval and its $2.8\sigma$ detection
floor reported once. The pooled rows are the analysis as first run ($4{,}317$ identities, 400
draws).
The HCOCO-only rows re-run the same estimator on the same records (post-hoc, $3{,}417$
identities, 2{,}000 draws, Supplement Table~S18 in Section~S14).

\begin{table}[htbp]
\centering
\small
\begin{tabular}{llrrr}
\toprule
Method & Rows & Step (MSE $\downarrow$) & 95\% CI & Detection limit \\
\midrule
DoveNet & pooled & $+16.36$ & $[-5.0, +40.8]$ & $33.8$ \\
  & HCOCO only & $+2.08$ & $[-8.5, +13.4]$ & $15.7$ \\
Harmonizer & pooled & $-62.31$ & $[-73.6, -49.4]$ & $17.5$ \\
  & HCOCO only & $-2.86$ & $[-7.0, +1.3]$ & $5.9$ \\
Editor (ours) & pooled & $-30.40$ & $[-47.9, -9.5]$ & $26.8$ \\
  & HCOCO only & $-2.32$ & $[-10.0, +5.8]$ & $11.0$ \\
\bottomrule
\end{tabular}
\caption{The matched exposure step under one estimator for all three methods, pooled and inside HCOCO. A
negative step means the shared composites score better once foreground ratio is matched (Supplement
Figure~S1). Detection limit: $2.8$ bootstrap standard errors.}
\label{tab:three}
\end{table}

\begin{figure}[t]
\centering
\includegraphics[width=\textwidth]{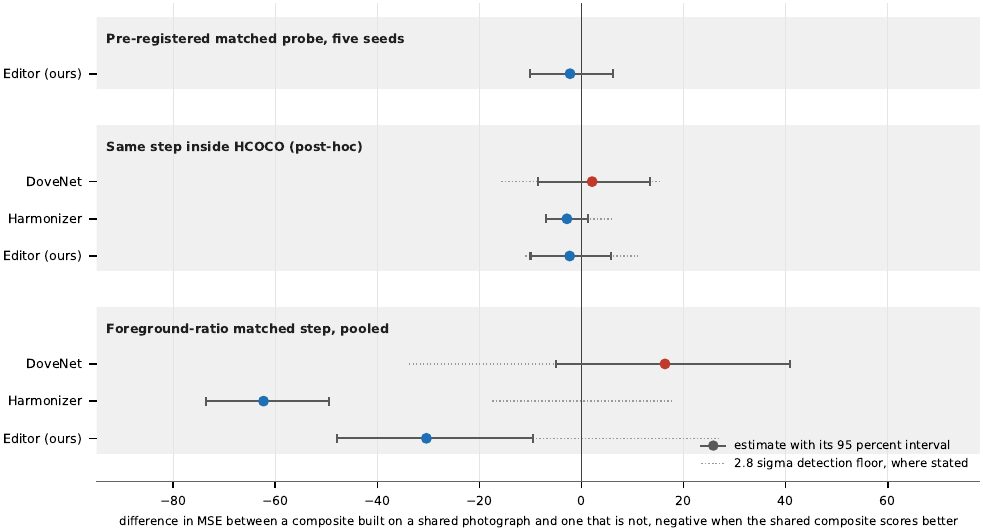}
\caption{The shared-composite step in whole-image MSE, so the intervals are comparable. It is large and
negative, pooled, for Harmonizer and for our editor. Inside HCOCO every interval includes zero, with
the detection floors drawn at $5.9$ to $15.7$ MSE.}
\label{fig:controls}
\end{figure}

Pooled, Harmonizer ($-62.31$ MSE) and our typed editor ($-30.40$) have intervals wholly below
zero.
DoveNet's step is positive and inside its $33.8$ MDE. The pooled estimator matches
foreground ratio but not sub-dataset: of the $1{,}822$ exposed composites $1{,}806$ are HCOCO and
16 HFlickr, against 812 unexposed HFlickr composites (HAdobe5k and Hday2night have none exposed), and the unexposed side ($5{,}582$) is 44 percent HCOCO (Supplement Table~S21 in Section~S14). Inside HCOCO, the only sub-dataset with
enough exposed composites for a within-sub-dataset contrast, every interval includes zero, and
Harmonizer's excludes any step below $-7.0$ against the pooled $-62.31$. The pooled-minus-HCOCO
difference is $95$ percent of Harmonizer's pooled step and $92$ percent of ours, with intervals
excluding zero (DoveNet's pooled step was not detected under this estimator).
 Figure~\ref{fig:controls} collects the MSE contrasts for the three methods scored end to end, on the shared
photographs on one axis, drawn against the detection floor of the design that produced it.

The same estimator on the three further checkpoints (post-hoc, Supplement Table~S18) gives
pooled and HCOCO-only MSE steps of $-42.49$ and $-1.40$ for PCT-Net and $-69.35$ and $-1.72$ for
DucoNet, the pattern of the table.

One direction does not hold. DoveNet's pooled matched step is positive ($+16.36$, inside its $33.8$
floor) while Harmonizer's and ours are negative and clear their floors ($-62.31$ and $-30.40$).
Inside HCOCO the three steps are $+2.08$, $-2.86$ and $-2.32$, all including zero. DiffHarmony goes
further and is detected inside HCOCO: $+50.23$ $[+41.0, +59.3]$ pooled and $+26.76$ $[+19.5,
+34.3]$ there, so shared composites score \emph{worse} for it. The reversal tracks the metric.
On fMSE, DoveNet is $+73.0$ worse on the excluded composites while Harmonizer is $-406.5$ better
and ours $-233.0$ better (Section~\ref{sec:inflation}). The same population moves an error metric in
opposite directions for different checkpoints, so the pooled step is a property of the checkpoint,
its training objective and the region each metric integrates, not a population-wide exposure
signal. 

In fMSE (not in the table), the post-hoc within-HCOCO residual is not used as evidence for the
main claim: under the stated multiplicity families, only Harmonizer and PCT-Net survive the
narrower family among the frozen-family methods, and neither survives the broader family
(Supplement Table~S23). The residual may reflect foreground size. The registered audit and the retrain that follows it
(Section~\ref{sec:e1observational}) use separate estimands and populations.

\noindent 

\subsection{Our own editor under the same protocol}
\label{sec:probe}

The third method is the one whose pipeline we control completely, so the model can be held
byte-identical while only the population changes: the five per-seed typed checkpoints of
Section~\ref{sec:compositing}, bound by SHA-256, are scored once over each population, inference
only, with no retraining, gate, threshold or selection, and the excluded list had not been scored
by this project before that pass. The control was pre-registered with its rule written down
before the numbers: if the foreground-matched gap collapses toward zero, the raw contrast is
population difficulty and no identity claim is licensed.

The raw contrast is large and in the same direction for every seed: the gap in mean absolute error
inside the object's region, the editor's own quality gate, is
positive in 5/5 seeds ($+1.954$ to $+2.938$). At $n = 5$ the smallest attainable two-sided sign-test value is $p = 0.0625$. MSE is
$3.06\times$ and fMSE $1.27\times$ lower on the excluded quarter. Within HCOCO, over eight
foreground-ratio quantile bins and a $2{,}000$-draw cluster bootstrap over $3{,}417$ target
identities, the matched excluded-minus-retained interval includes zero on all three metrics (MSE
$-2.25$ $[-10.09, +6.20]$, fMSE $-42.77$ $[-97.12, +15.56]$, PSNR $+0.04$ dB $[-0.22, +0.30]$).
The excluded composites are also easier before any model touches them (do-nothing reference and
the full table: Supplement Table~S3). The pre-committed rule applies: for our own checkpoints,
the target photograph is not separable from difficulty at this sample size. This population is
HCOCO only, and on it the statistic of Section~\ref{sec:three} reads the same for all three
methods (MSE steps $+2.08$, $-2.86$ and $-2.32$, intervals including zero, and ours agrees
with this probe's $-2.25$).

The arithmetic claim stands: an official-list score is a weighted average
over two populations that these methods score very differently, so official-list and
retained-list numbers are not interchangeable.

\subsection{The registered audit and the intervention}
\label{sec:e1observational}

In the pre-registered four-method fMSE audit (below its registered five-checkpoint roster
floor), the retained-versus-excluded interaction gate passed after target-photograph identity
bootstrap, foreground matching, and Holm correction. All six fMSE pair interactions in the frozen
DoveNet, Harmonizer, PCT-Net and DucoNet family were significant after Holm correction, and their intervals excluded zero under pooled foreground matching. The audit
compares the retained 5{,}581 composites with the excluded 1{,}823 (2{,}000 target-photograph identity
bootstrap replicates, seed 0, $\alpha = 0.05$, eight pooled foreground-ratio bins).

The registered audit reproduced the previously measured pair contrast: the DoveNet-minus-Harmonizer
fMSE gap widened by $+479.50$ on excluded versus retained composites (95\% CI $[403.73, 553.03]$ from the seed-0 2,000-draw run, $+504.60$ $[423.65, 583.84]$ foreground-matched; Supplement Table~S20 gives the same point estimate under its per-stratum reseeded run, $[+403.61, +557.22]$), against the pre-registered $+479$ point prediction. That prediction was computed from shifts already measured on the same two checkpoints
($+72.96$ and $-406.54$ fMSE), so the point value repeats an earlier measurement.
What the audit
adds is the clustered interval, the Holm family and the PCT-Net and DucoNet checkpoints.
Post-hoc, inside HCOCO alone, the interaction
is $-10.68$ $[-77.97, +56.26]$, so the pooled interaction is not detected within the one
sub-dataset where it can be estimated (Supplement Table~S20). Pairwise gaps depend on the
evaluation population: DoveNet-minus-Harmonizer is $+699.88$ on excluded and $+220.38$ on
retained composites.
Harmonizer-minus-DucoNet changes from $+47.16$ to $-11.53$, a descriptive
point-estimate reversal. 

\paragraph{Execution of the intervention.}
\label{sec:retrain}

Both evaluation difference-in-differences (DiD) intervals span zero and their design departed from
the registered execution, so neither the difference nor practical equivalence is established.  DoveNet's exposure DiD
was $-8.37$ fMSE (95\% CI
$[-41.61,+23.90]$, MDE $47.09$ fMSE under the $2.8\times$SE convention) and Harmonizer's $+8.23$
($[-37.78,+59.64]$, MDE $65.22$). Table~\ref{tab:retrain} keeps the fMSE between-arm change and the training-check
manipulation-check contrast separate from the audit and matched-step tables.

The registered design trains DoveNet and Harmonizer in two arms per method and three paired seeds,
with 63{,}226 training composites per arm including a matched 1{,}962-composite arm set (registered
budgets, epochs and steps in Supplement Section~S12). The two arms are the full-list arm, which keeps the $1{,}962$ exposed composites in training, and the reduced-list arm, which replaces them with a seeded, sub-dataset-matched sample of unexposed composites; both arm lists carry $63{,}226$ composites: a common $61{,}264$-composite unexposed base plus the arm's own $1{,}962$ composites, drawn from the $65{,}742$-row official training list. The estimand is exposed-minus-rest fMSE in the full-list arm minus the same contrast in the
reduced-list arm, stratified by subdataset and averaged over seed pairs.
On the test side, exposed comprises 1{,}874 composites and rest 5{,}530 retained composites after removing 51 exposed overlaps. Uncertainty uses 2{,}000 joint-bootstrap replicates over seed pairs and target
identities within subdataset, seed 0. The fixed equivalence margin is $(-25,+25)$ fMSE.
Neither interval lies inside that margin. Three seed pairs give coarse seed uncertainty. MDE is $2.8$ times
bootstrap SE, an approximate detection limit.

\begin{table}[htbp]
\centering
\small
\begin{tabular}{l p{4.3cm} r r r}
\toprule
Method & Estimand & fMSE contrast & 95\% joint-bootstrap CI & MDE \\
\midrule
DoveNet & Change between arms in the shared-composite advantage & $-8.37$ & $[-41.61,+23.90]$ & $47.09$ \\
Harmonizer & Change between arms in the shared-composite advantage & $+8.23$ & $[-37.78,+59.64]$ & $65.22$ \\
\midrule
DoveNet & Training check on the arms' own composites & $-308.75$ & $[-409.88,-234.13]$ & n/a \\
Harmonizer & Training check on the arms' own composites & $-56.51$ & $[-69.96,-41.08]$ & n/a \\
\bottomrule
\end{tabular}
\caption{The registered intervention and the training check, three seed pairs per method. The training-check
contrasts compare the full-list and reduced-list arms on the frozen 1{,}962-composite manipulation
population, and both exclude zero.}
\label{tab:retrain}
\end{table}

The training-check contrasts of Table~\ref{tab:retrain}, with all three seed pairs contributing for each
method, show that the two arms' checkpoints differ on composites that were in one arm's training
set and not the other's. They support that the training manipulation affected performance on its
specified population.  The training check was scored after the
pre-registration tag, a disclosed deviation, and does not alter the registered verdict
calculation. Provenance of the intervention and training-check values is detailed in Supplement Section~S9.

\paragraph{Sensitivity reads.}
The frozen audit gate passes under fMSE, MSE and PSNR (fMSE binding), with six, four and five of
six pairs Holm-significant and all six interaction directions in agreement after orienting PSNR
to the error direction.
The Harmonizer-DucoNet point-estimate reversal occurs under fMSE and MSE,
and PSNR has none. 

\subsection{External comparison on ccHarmony}
\label{sec:e2}

The same four checkpoints were evaluated zero-shot on the 1{,}180 ccHarmony test composites over
100 target photographs. At pairwise sign level, agreement with ccHarmony is 5 of 6 for the retained
list and 6 of 6 for the official list (the 7{,}404-composite union of retained and excluded
rows), so $\Delta_{\mathrm{agreement}}$ is $-1$ of 6.
Dropping either DucoNet or Harmonizer moves
the leave-one-method-out (LOMO) sign to zero.
The retained-versus-ccHarmony Kendall tau is
$0.6667$ $[0.6667,1.0]$. The pre-registered prediction that retained-list rankings agree
with zero-shot ccHarmony behavior at least as well as official-list rankings, is mixed at sign
level and is not supported at the observed point estimates. The comparison is descriptive only:
it had no gate or success threshold, it carries no power claim, and it does not show external
validity of the retained list (quantities in Supplement Table~S8, with signs on point gaps).

For a reader of published numbers, the pooled differences in this section are arithmetic over a mixture the benchmark does not disclose.
The reported performance difference between two public methods is $2.3$ times larger on the official list than on the retained list in point estimates. Inside HCOCO the same gap is smaller on the official list than on the retained list, so the shared composites narrow the better method's advantage there (Section~\ref{sec:gaps}). The DoveNet-Harmonizer gap is positive in point estimate in every population we scored; its interval excludes zero in all of them except the pooled retained list.
The
Harmonizer-minus-DucoNet gap does change sign in the descriptive point estimates
(Section~\ref{sec:e1observational}). A number on this benchmark is averaged over a population whose composition decides part of it, and
a method that benefited from shared photographs would be rewarded for that rather than for harmonization.

\section{The retained list}
\label{sec:release}

\subsection{Adopting the retained list}
\label{sec:ships}

The retained list is a replacement test list of 5{,}581 composites, kept beside the unmodified
official list, so the 1{,}823 excluded composites are the set difference of the two and the training
list is unchanged. In the project's development repository it comes with the
identity matcher and its manifest, regression tests, and a re-derivation driver with a pin
registry.
The artifact repository shipped with this paper carries the matcher, the lists and the
CPU checks, not the driver. The driver re-scores a public baseline on the official, retained or excluded list from pinned
upstream commits and checkpoint digests, refuses to run against an unpinned or mismatched checkpoint, and checks the
staged tree against both lists before scoring (a check that caught a real gap in our own
staging). 

Adopting the list means pointing the evaluation at the retained list instead of the official list and reporting the retained number beside the official one rather than in place of it. On the
same pinned checkpoints scored by the same driver, retained MSE is worse for both methods (DoveNet
$122.75 \rightarrow 140.18$, Harmonizer $107.08 \rightarrow 133.42$) and the pooled method-to-method MSE gap shrinks $2.3\times$ in point estimates ($+15.67 \rightarrow +6.76$; the ratio interval $[-13.42, +15.16]$ includes 1 and the retained gap interval $[-5.20, +18.74]$ includes zero; Section~\ref{sec:gaps}).
On fMSE the two methods move in opposite
directions, the foreground-size difference of Section~\ref{sec:inflation}. The columns are not
interchangeable and the difference is not a uniform correction (full table with fMSE, PSNR and
changes: Supplement Table~S13).

\subsection{Evaluation-hygiene rules}
\label{sec:hygiene}

The audit and a compositing-metric defect we found in our own trainers lead to five short rules.
\textbf{H1. Match identities, not files.} File-level deduplication does not detect sharing of
the real photograph a composite is built from.
Match on the identity that generated the sample,
fail closed, and audit the perceptual residue rather than assuming it is zero
(Section~\ref{sec:perceptual}). \textbf{H2. Correct the evaluation list, not the training list.}
Where leaked training images cannot be removed, correcting the test side is the cheaper
remedy~\cite{ramos2025leakage}: it keeps every published checkpoint eligible for rescoring, while a
correction that requires retraining is costlier to adopt. \textbf{H3. Score a composited editor
after compositing.} Scoring the pre-composite tensor describes a different system with no
locality guarantee.
It moved one of our reported metrics by 8.2 dB in the wrong direction
(Section~\ref{sec:asymmetry}). \textbf{H4. Disclose the compositing-metric asymmetry.}
Whole-image MSE and PSNR mechanically reward a composited editor, since every unedited pixel is
exactly right, while a full-image baseline pays for every pixel.
A benchmark mixing the two
should say so and lead harmonization-quality comparisons with fMSE, as we do, and on it our editor is not the lead row.
\textbf{H5. Declare wear and score final numbers once, on a list-defined population,} after
reporting how much of the test split the development run already touched and pre-committing the
remedy.

\section{The third method and its certificate}
\label{sec:compositing}

\label{sec:asymmetry}
The results above need one method whose training data, compositing rule and evaluation convention we
can inspect end to end, because the published baselines do not all expose them. We therefore train a
small harmonization network of our own and freeze it as a third checkpoint. It is a control in this
study rather than a harmonization-quality claim.

The control is useful here for one reason: its channel from outside the edited region to the pixels
written inside it is bounded, and the bound can be computed rather than assumed. Written out in
words, the network admits outside information through two paths. The first is a set of six global
colour statistics of the whole frame, and the second is a pooled feature of the frame that we disable
in the reported configuration. The output inside the scored region is a sum of a local path that
reads only the trusted window, a physical term driven by those six statistics, and a semantic term
driven by the pooled feature. For the architecture we ship, the worst-case change that any
perturbation outside the trusted window can induce inside the scored region is bounded by a product
of per-layer operator norms, and with the pooled path disabled that bound is exactly zero for the
semantic term. The bound is proved for the stated architecture in Supplement Section~S15, with an
outline in Section~S11, and it is checked in the repository three ways: closed-form re-derivation,
regression tests at odd resolutions under outside-supported perturbations, and a negative control
that can fail. Those checks are consistency checks, not a proof. The bound is loose by about a factor
of thirteen, and a held-out projected-gradient search over 500 attacks never exceeded it
(Supplement Section~S6). 
Two properties should be kept apart: the composite returns the source pixel exactly where
the effective write support is zero, which is a property of the write set and not of the bound, and
the bound above is a property of the influence path, not of the write set. We call the control \emph{typed} because its outside
channels are named and separated, physical and semantic, each with its own bound.

\textbf{Metric choice.} Our editor writes inside a
feathered region and composites elsewhere, so every unedited pixel is exactly right, and whole-image
MSE and PSNR reward that convention mechanically while a baseline that emits a full image pays for
every pixel. Only composited values computed after the scoring fix of 2026-08-21 appear in this
paper. Before that fix our own trainers reported the raw source-plus-residual on a 128-composite
check, which described an editor with no such guarantee: MSE $572.62$ and PSNR $21.83$\,dB against
doing nothing at $325.72$ and $30.01$\,dB, $8.2$\,dB worse, with mean background leakage of
$15.685/255$. After compositing, the same checkpoint reads MSE $267.87$, PSNR $30.40$\,dB and fMSE
$1550.55$, and the mean absolute error inside the object's region, the pass/fail gate of the editor's
own training check, moves only from $29.592$ to $29.662$, so no verdict changed. Every pre-fix value is invalid and is cited only here. Every table in this paper therefore
states its metric, and rule H4 of Section~\ref{sec:hygiene} follows: the population arithmetic is
reported in whole-image MSE, because the two populations are the object of study, and the comparisons
of harmonization quality are reported in fMSE.

\label{sec:metrics}
\textbf{Placement of the third method.} On the standard protocol, every method scored on the
identical retained list, the editor is level with DoveNet on fMSE ($1104.82$ against $1114.71$) and
behind Harmonizer ($894.34$), and its error inside the object's region falls from the no-op source's $29.64$ to $25.01$
$[24.45, 25.56]$. That places a demonstration-scale network in the benchmark.
We make no quality
comparison with recent generative harmonizers~\cite{zhou2024diffharmony,zhang2025r2r}. The placement
table is Supplement Table~S10. The paired ablation that measures what the certified channel is worth
against a capacity-matched floor, and the price paid for the certificate, are Supplement Table~S9.
Of the six pre-registered predictions about the editor, one was falsified and is kept: P2, that
$K$-draw floors would remove at least $0.4$ L1 of the apparent effect, came in at $+0.023$, so that
selection route does not explain the effect.

\section{Related work}
\label{sec:related}

\paragraph{Harmonization and its protocol.} iHarmony4 and DoveNet~\cite{cong2020dovenet} define the
task and the lists we audit. The checkpoints we score are DoveNet, Harmonizer~\cite{ke2022harmonizer},
PCT-Net~\cite{guerreiro2023pctnet}, DucoNet~\cite{tan2023duconet} and, descriptively,
DiffHarmony~\cite{zhou2024diffharmony}.
ccHarmony~\cite{niu2023ccharmony} is the external test
set of Section~\ref{sec:e2}. R2R~\cite{zhang2025r2r} is a recent latent-diffusion harmonizer. We
did not survey which published methods evaluate on identity-disjoint lists. Ramos et
al.~\cite{ramos2025leakage} detect leakage between training data and evaluation benchmarks with
image retrieval, and recommend removing leaked items from training, or failing that from the test
side.
We follow the test-side remedy here. In iHarmony4 the shared item is the real photograph
composites are built from, which appears under different file names.
We did not test
retrieval-based detectors on this case.

\paragraph{Audits of train-test overlap.} Overlap between evaluation and training data has been
audited in classification, where near-duplicates between the two sides inflate measured accuracy:
Barz and Denzler remove the test duplicates of CIFAR and re-score published architectures
~\cite{barz2020purging}, and Recht et al. build fresh test sets to separate a real generalisation gap
from tuning to the old one~\cite{recht2019generalize}. Auditing practice extends past duplicates:
Northcutt et al. relabel test sets and show that rankings hold on the whole set and flip on the
correctable part as the error rate rises~\cite{northcutt2021label}, Gema et al. annotate errors in a
language benchmark and release a corrected subset~\cite{gema2025mmlu}, Lewis et al. stratify
question-answering test sets by overlap with training~\cite{lewis2021overlap}, Sturm documents the
faults of a music-genre dataset and their effect on evaluation~\cite{sturm2013gtzan}, Allamanis
measures duplication between the training and evaluation splits of code models~\cite{allamanis2019duplication},
and Engstrom et al. show that a gap between two populations can come from how the second population was
constructed rather than from the models~\cite{engstrom2020replication}. Kapoor and Narayanan place
nonindependence between training and test samples in a taxonomy of leakage, and note that file-level
checks cannot see it~\cite{kapoor2023leakage}. What none of them does is audit the construction unit of
a composited benchmark, where the shared element is a photograph rather than a file.

\paragraph{Write-side locality.} Blended Diffusion~\cite{avrahami2022blended},
DiffEdit~\cite{couairon2023diffedit}, Watch Your Steps~\cite{mirzaei2024watch},
ZONE~\cite{li2024zone}, BrushNet~\cite{ju2024brushnet} and EditMGT~\cite{chow2026editmgt} localize
edits and preserve the background by design, and PIE-Bench~\cite{ju2024pnp} and
GIE-Bench~\cite{qian2025giebench} measure background preservation (GIE-Bench reports that
GPT-Image-1 often over-modifies irrelevant regions). In the sources we read, these preservation
methods bound or measure what changes rather than which outside pixels caused the change.
Concurrent 2026 work addresses constrained or decomposed editing and preservation behavior:
constrained edit fields and spatial edit responsibility~\cite{kang2026cef}, attention-scoped
spatial guidance~\cite{li2026asg}, decomposed efficacy and locality rewards~\cite{chen2026arro},
and a four-axis verifiable editing benchmark~\cite{wang2026veriedit}.
None supplies the
fixed-mask, exterior-supported worst-case sensitivity certificate instantiated here.
Lukoianov et al.~\cite{lukoianov2025locality} derive denoiser locality from data statistics and
therefore do analyse which input neighborhoods a denoiser reads.
Hu et al.~\cite{hu2026manifold}
give a conditional, first-order leakage bound for diffusion editors under Lipschitz assumptions and
label its tables illustrative. Neither supplies the fixed-mask, exterior-supported, worst-case bound
on the target output instantiated here.
Hu's bound is conditional rather than an instantiated
constant, and both sources do address how an input region influences the output.

\paragraph{Certified bounds and attacks.} We reuse FFT layer norms~\cite{sedghi2019singular} and norm-preserving
activations~\cite{anil2019sorting}, developed in the Lipschitz-network literature.
LipNeXt's
evaluated guarantees concern classification~\cite{hu2026lipnext}, and
Cert-LAS~\cite{qi2026certlas} uses smoothing for model-ownership verification of text-to-image
diffusion, not for an editing-context certificate. In the sources we read, none certifies an
out-of-region-input to in-region-output editing map. Attack-side work is
owner-side and whole-image (PhotoGuard~\cite{salman2023photoguard},
immunization~\cite{lee2026immunization}), targets text-reading multimodal models (image-based
prompt injection~\cite{nagaraja2025promptinjection}), or triggers moderation refusals in
commercial editors (MIRAGE~\cite{nasery2026mirage}).
A different method also named
MIRAGE~\cite{liu2026mirage} performs localized multi-instance editing.

\section{Limitations}
\label{sec:limitations}

\textbf{What the study covers.} The population contrasts rest on DoveNet, Harmonizer and our own
editor, and the five-checkpoint roster adds PCT-Net and DucoNet, which follow the pooled pattern, and
DiffHarmony, which reverses it. That is a study of the checkpoints we scored, not a census of the
field. The added evidence is demonstration-scale: both evaluation difference-in-differences intervals
span zero, with minimum detectable effects of $47.09$ and $65.22$ fMSE, so they are not interpretable
as a null.
The training-check intervals exclude zero on the training-manipulation population, and the
post-tag collection of that check is a deviation rather than a change to its reading. HCOCO dominates
the stratified contrast and seed uncertainty is coarse.

\textbf{Departures in the intervention.}
The design pre-registered twelve runs with both arms of a seed pair trained in one driver process and
the step budget enforced by the driver. Six of the twelve arms were interrupted and resumed.
Six of
the eight resume events re-applied the run seed instead of continuing the random stream, and in three arms (reduced-list DoveNet seeds 0 and 2 and full-list DoveNet seed 1) the resumed loop
replayed a partial epoch so the weights received about $2.1$, $2.1$ and $1.0$ percent more updates
than the budget while the step counters still read the budget. The DoveNet
trainer saves weights only, so optimizer state was not carried. Because the arms of a pair came from
separate invocations, the driver's automatic equal-step check did not run.
Matching seeds, step
totals, row counts and weight hashes do not show that a run was uninterrupted, or that the arms differ
only in training exposure. Harmonizer seeds 1 and 2 were trained on different GPUs under different
torch releases, and launch commits differ across arms although the pinned scripts are byte-identical.
We therefore use the saved difference-in-differences values as descriptive results with their
intervals and detection limits, and draw no causal, null or equivalence conclusion from them.
Per-arm run records are in the project repository. The retrained DoveNet uses instance normalization
where its leaderboard checkpoint uses batch normalization, so absolute fMSE is not comparable across
the two.

\textbf{Composition as an alternative explanation.} The pooled matched step matches foreground
ratio, not sub-dataset. Only HCOCO holds enough exposed composites for a within-sub-dataset contrast
(HAdobe5k and Hday2night hold none, HFlickr holds 16 against a minimum of 30), so most of each pooled
step is a contrast across sub-datasets, and inside HCOCO the MSE steps are not detected for any of the
three methods, at floors of $5.9$ to $15.7$. That absence of detection is not a showing of no
difference. Exposure inside HCOCO is not randomized, the stratified intervals are uncorrected for
multiplicity, dose slopes beyond $k = 1$ did not exclude zero under the stated specifications, and the
remaining post-hoc fMSE steps may reflect foreground size.

\textbf{Boundaries of the audit and of the third method.} The matcher is path-identity based, so
near-duplicates, same-scene frames, crops and resizes carry different identity strings.
The measured residue is four confirmed pairs plus one further whole-matrix candidate, five target photographs in all, $0.17$ percent of the $2{,}917$ retained backgrounds ($0.14$ percent for the four confirmed pairs alone), with wider-radius candidates disclosed as unadjudicated, so the retained list is not a clean split, and the cross-benchmark scan is post-hoc and
name-level. The external ccHarmony comparison is mixed descriptive agreement, with that prediction unsupported at
the observed point estimates and a Harmonizer-DucoNet reversal under fMSE and MSE. About $92$ percent
of the retained test subset had already been scored by the development run's own test-side subsets, so
the full-list placement carries no fresh-test claim. The third method is a small CNN whose real-arithmetic certificate is checked by 500 $\Omega$-confined
PGD searches that never exceeded the numerical bound.
The bound carries about $13\times$ slack and
sits below the edit magnitude for two of five seeds. The tightness figure is an attack-dependent
lower bound, and promoted weights are not bound to a pre-runtime lock. Its trained regularization penalty is not a
certificate axis, being a sum of reshaped-matrix spectral norms against a product of FFT norms
measured $27.7$ times apart, and its FFT synthesis norm is exact only at the computed resolution,
with the transfer tensor memory-bound at $5.37$\,GB for $n = 1024$. Shadow
removal on ISTD~\cite{wang2018istd} is a concluded negative whose floor-minus-typed gap was $+0.619$
$[-2.99, +4.23]$ and is not re-run here. Trained numbers from before the determinism fix carry a $1.5$ L1 run-to-run noise floor
(mean $\lvert \Delta \rvert = 1.47$) and are never aggregated with deterministic-era numbers.
\section{Reproducibility and provenance}

\noindent The observational analyses in this paper read saved per-composite records: the pinned
re-derivation run, which stores per-composite records with foreground ratios for DoveNet and
Harmonizer on all three lists and whose identifier is in Supplement~S7, and
the once-scored five-seed inference records of our own editor.
\label{sec:repro}

\textbf{Provenance.} The numbers in this paper are read from saved run records: per-composite
score files, per-seed checkpoints bound by SHA-256, and summary files. Unless the text calls a
value a registered design value (the retrain budgets of Supplement Section~S12, the $+479$
prediction), it is a measured value. Nothing was changed in any source record or experiment for
this version of the paper. No new training run was made for it, and the scorings added after the audit
are inference passes over the recorded lists.

\textbf{What regenerates.} The cross-method contrasts, the dose statistic, the matched exposure
steps and their post-hoc stratification, the near-duplicate audit and Supplement Table~S10
regenerate on CPU from the re-derivation registry and per-composite records, using scripts
tracked in the project repository. The table driver recomputes aggregates from per-sample rows or
per-seed blocks and refuses to print a row it cannot re-derive. The two distributed lists carry recorded content hashes in Supplement~S10; the staged tree, the earlier retained-only staging tree used by the gate runs, and the code fingerprint of the gate runs are recorded in the project repository. The split-identity counts
of Section~\ref{sec:leak} are recomputed on CPU from the distributed lists by a short script. A second, independent pass
regenerated the difference-in-differences and training-check estimates, bootstrap endpoints,
standard errors and detection limits from the saved rows exactly (Supplement Section~S9).

The retained list, the identity matcher and the CPU verification checks are released under the MIT license at
\url{https://github.com/Buljic/only-means-only}. The retrain code and the baseline scoring code are not, so the retrain estimates and
the population tables cannot be recomputed from the distributed artefact alone: the per-composite
records they read are distributed, and the estimates are recorded in the tables.

Generative AI tools were used for standard code and language editing.

\section{Conclusion}
\label{sec:conclusion}

The official iHarmony4 test list has 1{,}399 target photographs in common with the train list, so
24.6 percent of its composites are built on a photograph the training list also uses. Pooled, that quarter scores $2$
to $5$ times lower in MSE for the three methods we score end to end, and the gap between two
published methods is $2.3\times$ larger on the official list in point estimates.
Inside HCOCO the factors are $1.47$ to $1.61$, and the matched MSE steps are not detected (Harmonizer $-2.86$, ours
$-2.32$, DoveNet $+2.08$) against pooled $-62.31$, $-30.40$ and $+16.36$. PCT-Net and DucoNet show the same pattern, with retained-to-excluded ratios of
$4.68$ and $5.73$, and DiffHarmony scores worse on the shared composites in MSE while the foreground-region metric moves the other way (Supplement Table~S19). We report the
identity sharing, a retained test subset, a whole-matrix near-duplicate audit, and five hygiene
rules. The study does not identify how much of any score difference is caused by the sharing itself.
Official and retained-list scores should be reported as distinct evaluation
populations.

\footnotesize
\bibliographystyle{unsrt}
\bibliography{references}

\begin{thebibliography}{10}

\bibitem{cong2020iharmony4}
Wenyan Cong, Li~Niu, Jianfu Zhang, Liu Liu, Zhixin Ling, Weiyuan Li, and Liqing
  Zhang.
\newblock Image harmonization dataset {iHarmony4}: Generate synthetic composite
  images for realistic image harmonization, 2019.
\newblock arXiv:1908.10526; Section 4 states the target-image separation
  promise.

\bibitem{zhou2024diffharmony}
Pengfei Zhou, Fangxiang Feng, and Xiaojie Wang.
\newblock {DiffHarmony}: Latent diffusion model meets image harmonization.
\newblock In {\em ACM International Conference on Multimedia Retrieval (ICMR)},
  2024.
\newblock arXiv:2404.06139.

\bibitem{barz2020purging}
Bj{\"o}rn Barz and Joachim Denzler.
\newblock Do we train on test data? purging {CIFAR} of near-duplicates.
\newblock {\em Journal of Imaging}, 6(6):41, 2020.
\newblock arXiv:1902.00423.

\bibitem{recht2019generalize}
Benjamin Recht, Rebecca Roelofs, Ludwig Schmidt, and Vaishaal Shankar.
\newblock Do {ImageNet} classifiers generalize to {ImageNet}?
\newblock In {\em International Conference on Machine Learning (ICML)}, 2019.
\newblock arXiv:1902.10811.

\bibitem{kapoor2023leakage}
Sayash Kapoor and Arindam Narayanan.
\newblock Leakage and the reproducibility crisis in machine-learning-based
  science.
\newblock {\em Patterns}, 2023.
\newblock arXiv:2207.07048.

\bibitem{niu2023ccharmony}
Li~Niu, Linfeng Tan, Xinhao Tao, Junyan Cao, Fengjun Guo, Teng Long, and Liqing
  Zhang.
\newblock Deep image harmonization with globally guided feature transformation
  and relation distillation.
\newblock In {\em IEEE/CVF International Conference on Computer Vision (ICCV)},
  2023.
\newblock arXiv:2308.00356; introduces the ccHarmony dataset.

\bibitem{hong2022desoba}
Yan Hong, Li~Niu, Jianfu Zhang, and Liqing Zhang.
\newblock Shadow generation for composite image in real-world scenes.
\newblock In {\em AAAI Conference on Artificial Intelligence}, 2022.
\newblock arXiv:2104.10338; introduces the DESOBA dataset.

\bibitem{guerreiro2023pctnet}
Julian Jorge~Andrade Guerreiro, Mitsuru Nakazawa, and Bj{\"o}rn Stenger.
\newblock {PCT-Net}: Full resolution image harmonization using pixel-wise color
  transformations.
\newblock In {\em IEEE/CVF Conference on Computer Vision and Pattern
  Recognition (CVPR)}, pages 5917--5926, 2023.

\bibitem{tan2023duconet}
Linfeng Tan, Jiangtong Li, Li~Niu, and Liqing Zhang.
\newblock Deep image harmonization in dual color spaces.
\newblock In {\em ACM International Conference on Multimedia (ACM MM)}, 2023.
\newblock arXiv:2308.02813.

\bibitem{ramos2025leakage}
Patrick Ramos, Ryan Ramos, and Noa Garcia.
\newblock Data leakage in visual datasets.
\newblock In {\em IEEE/CVF International Conference on Computer Vision
  Workshops (ICCVW)}, pages 6368--6378, 2025.
\newblock arXiv:2508.17416.

\bibitem{zhang2025r2r}
Zhiqiu Zhang, Dongqi Fan, Mingjie Wang, Qiang Tang, Jian Yang, and Zili Yi.
\newblock Region-to-region: Enhancing generative image harmonization with
  adaptive regional injection, 2025.
\newblock arXiv:2508.09746.

\bibitem{cong2020dovenet}
Wenyan Cong, Jianfu Zhang, Li~Niu, Liu Liu, Zhixin Ling, Weiyuan Li, and Liqing
  Zhang.
\newblock {DoveNet}: Deep image harmonization via domain verification.
\newblock In {\em IEEE/CVF Conference on Computer Vision and Pattern
  Recognition (CVPR)}, 2020.
\newblock arXiv:1911.13239.

\bibitem{ke2022harmonizer}
Zhanghan Ke, Chunyi Sun, Lei Zhu, Ke~Xu, and Rynson W.~H. Lau.
\newblock Harmonizer: Learning to perform white-box image and video
  harmonization.
\newblock In {\em European Conference on Computer Vision (ECCV)}, 2022.
\newblock arXiv:2207.01322.

\bibitem{northcutt2021label}
Curtis~G. Northcutt, Anish Athalye, and Jonas Mueller.
\newblock Pervasive label errors in test sets destabilize machine learning
  benchmarks.
\newblock In {\em Advances in Neural Information Processing Systems (NeurIPS)},
  2021.
\newblock arXiv:2103.14749.

\bibitem{gema2025mmlu}
Aryo~Pradipta Gema, Joshua Ong~Jun Leang, Giwon Hong, Alessio Devoto, Alberto
  C.~S. Mancino, Rohit Saxena, Xuanli He, Yilun Zhao, Xiaotang Yang, et~al.
\newblock Are we done with {MMLU}?
\newblock In {\em Conference of the Nations of the Americas Chapter of the ACL
  (NAACL)}, 2025.
\newblock arXiv:2406.04127.

\bibitem{lewis2021overlap}
Patrick Lewis, Pontus Stenetorp, and Sebastian Riedel.
\newblock Question and answer test-train overlap in open-domain question
  answering datasets.
\newblock In {\em Conference of the European Chapter of the ACL (EACL)}, 2021.
\newblock arXiv:2008.02637.

\bibitem{sturm2013gtzan}
Bob~L. Sturm.
\newblock The {GTZAN} dataset: Its contents, its faults, their effects on
  evaluation, and its future use, 2013.
\newblock arXiv:1306.1461.

\bibitem{allamanis2019duplication}
Miltiadis Allamanis.
\newblock The adverse effects of code duplication in machine learning models of
  code.
\newblock In {\em ACM SIGPLAN International Symposium on New Ideas, New
  Paradigms, and Reflections on Programming and Software (Onward!)}, 2019.
\newblock arXiv:1812.06469.

\bibitem{engstrom2020replication}
Logan Engstrom, Andrew Ilyas, Shibani Santurkar, Dimitris Tsipras, Jacob
  Steinhardt, and Aleksander Madry.
\newblock Identifying statistical bias in dataset replication.
\newblock In {\em International Conference on Machine Learning (ICML)}, 2020.
\newblock arXiv:2005.09619.

\bibitem{avrahami2022blended}
Omri Avrahami, Dani Lischinski, and Ohad Fried.
\newblock Blended diffusion for text-driven editing of natural images.
\newblock In {\em IEEE/CVF Conference on Computer Vision and Pattern
  Recognition (CVPR)}, 2022.
\newblock arXiv:2111.14818.

\bibitem{couairon2023diffedit}
Guillaume Couairon, Jakob Verbeek, Holger Schwenk, and Matthieu Cord.
\newblock {DiffEdit}: Diffusion-based semantic image editing with mask
  guidance.
\newblock In {\em International Conference on Learning Representations (ICLR)},
  2023.
\newblock arXiv:2210.11427.

\bibitem{mirzaei2024watch}
Ashkan Mirzaei, Tristan Aumentado-Armstrong, Marcus~A. Brubaker, Jonathan
  Kelly, Alex Levinshtein, Konstantinos~G. Derpanis, and Igor Gilitschenski.
\newblock Watch your steps: Local image and scene editing by text instructions.
\newblock In {\em European Conference on Computer Vision (ECCV)}, 2024.
\newblock arXiv:2308.08947.

\bibitem{li2024zone}
Shanglin Li, Bohan Zeng, Yutang Feng, Sicheng Gao, Xuhui Liu, Jiaming Liu,
  Li~Lin, Xu~Tang, Yao Hu, Jianzhuang Liu, and Baochang Zhang.
\newblock {ZONE}: Zero-shot instruction-guided local editing.
\newblock In {\em IEEE/CVF Conference on Computer Vision and Pattern
  Recognition (CVPR)}, 2024.
\newblock arXiv:2312.16794.

\bibitem{ju2024brushnet}
Xuan Ju, Xian Liu, Xintao Wang, Yuxuan Bian, Ying Shan, and Qiang Xu.
\newblock {BrushNet}: A plug-and-play image inpainting model with decomposed
  dual-branch diffusion.
\newblock In {\em European Conference on Computer Vision (ECCV)}, 2024.
\newblock arXiv:2403.06976.

\bibitem{chow2026editmgt}
Wei Chow, Linfeng Li, Lingdong Kong, Zefeng Li, Qi~Xu, Hang Song, Tian Ye, Xian
  Wang, Jinbin Bai, Shilin Xu, Xiangtai Li, Junting Pan, Shaoteng Liu, Ran
  Zhou, Tianshu Yang, and Songhua Liu.
\newblock {EditMGT}: Unleashing potentials of masked generative transformers in
  image editing.
\newblock In {\em IEEE/CVF Conference on Computer Vision and Pattern
  Recognition (CVPR)}, 2026.
\newblock arXiv:2512.11715.

\bibitem{ju2024pnp}
Xuan Ju, Ailing Zeng, Yuxuan Bian, Shaoteng Liu, and Qiang Xu.
\newblock {PnP} inversion: Boosting diffusion-based editing with 3 lines of
  code.
\newblock In {\em International Conference on Learning Representations (ICLR)},
  2024.
\newblock Introduces PIE-Bench; arXiv:2310.01506 (as Direct Inversion).

\bibitem{qian2025giebench}
Yusu Qian, Jiasen Lu, Tsu-Jui Fu, Xinze Wang, Chen Chen, Yinfei Yang, Wenze Hu,
  and Zhe Gan.
\newblock {GIE-Bench}: Towards grounded evaluation for text-guided image
  editing, 2025.
\newblock arXiv:2505.11493.

\bibitem{kang2026cef}
Jingxuan Kang, Yinsong Wang, Che Liu, and Chen Qin.
\newblock Constrained edit fields for training-free flow editing, 2026.
\newblock arXiv:2609.33735.

\bibitem{li2026asg}
Zeyan Li, Wei Zhou, Hadi Amirpour, Minghao Zou, Panqi Yang, and Jianfeng Xu.
\newblock Attention-scoped guidance: Training-free spatial control for image
  editing, 2026.
\newblock arXiv:2609.37492.

\bibitem{chen2026arro}
Shuo Chen, Fengming Huang, Yu~Yao, Mingming Gong, and Tongliang Liu.
\newblock Scalable minimal-change learning for controllable image editing,
  2026.
\newblock arXiv:2610.06021; the arXiv comment lists NeurIPS 2026.

\bibitem{wang2026veriedit}
Muyao Wang, Chen Zhu, Shiqi Yang, DongHyun Hwan, and Hideki Nakayama.
\newblock Beyond plausibility: Verifiable fine-grained image editing on
  structured assets, 2026.
\newblock arXiv:2610.04381.

\bibitem{lukoianov2025locality}
Artem Lukoianov, Chenyang Yuan, Justin Solomon, and Vincent Sitzmann.
\newblock Locality in image diffusion models emerges from data statistics,
  2025.
\newblock arXiv:2509.09672.

\bibitem{hu2026manifold}
Yi~Hu, Leying Yi, Emily Davis, and Finn Carter.
\newblock Editing on the generative manifold: A theoretical and empirical study
  of general diffusion-based image editing trade-offs, 2026.
\newblock arXiv:2603.29736.

\bibitem{sedghi2019singular}
Hanie Sedghi, Vineet Gupta, and Philip~M. Long.
\newblock The singular values of convolutional layers.
\newblock In {\em International Conference on Learning Representations (ICLR)},
  2019.
\newblock arXiv:1805.10408.

\bibitem{anil2019sorting}
Cem Anil, James Lucas, and Roger Grosse.
\newblock Sorting out {L}ipschitz function approximation.
\newblock In {\em International Conference on Machine Learning (ICML)}, 2019.
\newblock arXiv:1811.05381.

\bibitem{hu2026lipnext}
Kai Hu, Haoqi Hu, and Matt Fredrikson.
\newblock {LipNeXt}: Scaling up {L}ipschitz-based certified robustness to
  billion-parameter models.
\newblock In {\em International Conference on Learning Representations (ICLR)},
  2026.
\newblock arXiv:2601.18513.

\bibitem{qi2026certlas}
Leyi Qi, Yiming Li, Siyuan Liang, Zhengzhong Tu, and Dacheng Tao.
\newblock {Cert-LAS}: Toward certified model ownership verification for
  text-to-image diffusion models via layer-adaptive smoothing.
\newblock In {\em International Conference on Machine Learning (ICML)}, 2026.
\newblock arXiv:2605.29809.

\bibitem{salman2023photoguard}
Hadi Salman, Alaa Khaddaj, Guillaume Leclerc, Andrew Ilyas, and Aleksander
  Madry.
\newblock Raising the cost of malicious {AI}-powered image editing.
\newblock In {\em International Conference on Machine Learning (ICML)}, volume
  202 of {\em PMLR}, pages 29894--29918, 2023.

\bibitem{lee2026immunization}
Chanhui Lee, Donggyu Choi, Seunghyun Shin, Hae-Gon Jeon, and Jeany Son.
\newblock Universal image immunization against diffusion-based image editing
  via semantic injection, 2026.
\newblock arXiv:2602.14679; the arXiv comment lists ECCV 2026.

\bibitem{nagaraja2025promptinjection}
Neha Nagaraja, Lan Zhang, Zhilong Wang, Bo~Zhang, and Pawan Patil.
\newblock Image-based prompt injection: Hijacking multimodal {LLMs} through
  visually embedded adversarial instructions.
\newblock In {\em International Conference on Foundation and Large Language
  Models (FLLM)}, pages 916--922, 2025.
\newblock arXiv:2603.03637.

\bibitem{nasery2026mirage}
Anshul Nasery, Ramnath Kumar, Cho-Jui Hsieh, and Sewoong Oh.
\newblock {MIRAGE}: Protecting against malicious image editing via false
  moderation, 2026.
\newblock arXiv:2606.26199.

\bibitem{liu2026mirage}
Ziqian Liu and Stephan Alaniz.
\newblock {MIRAGE}: Benchmarking and aligning multi-instance image editing,
  2026.
\newblock arXiv:2604.05180.

\bibitem{wang2018istd}
Jifeng Wang, Xiang Li, and Jian Yang.
\newblock Stacked conditional generative adversarial networks for jointly
  learning shadow detection and shadow removal.
\newblock In {\em IEEE/CVF Conference on Computer Vision and Pattern
  Recognition (CVPR)}, 2018.
\newblock arXiv:1712.02478; introduces the ISTD dataset.

\end{thebibliography}

\end{document}

\section*{Supplementary Material}

\noindent This is the supplement to ``Only Means Only: what a harmonization benchmark score averages
over''. Main-text numbers are cited as ``main Section N'' or ``main Table N'', and the figures and
tables here are numbered S1 onwards. 
Values are measured from saved run records unless the text calls them registered design values, the
budgets and predictions fixed before a measurement.

\section{Graded exposure}

The dose $k$ is the number of train composites built from the same target photograph as a test
composite.
The dose distribution over the 7{,}404 official test composites is $k = 0$: 5{,}582, $k = 1$:
1{,}024, $k = 2$: 443, $k = 3$: 161, $k = 4$: 103, $k = 5$: 45, $k = 6$: 12, $k = 7$: 25,
$k = 8$: 8, $k = 9$: 1. The $k = 0$ count is 5{,}582 rather than 5{,}581 because one of the
1{,}823 excluded composites was excluded as a decoded-pixel duplicate (its target is
pixel-identical to a train target under a \emph{different} filename), so its identity string
never occurs in the train list and the dose counter correctly assigns it $k = 0$. The remaining
$1{,}822$ exclusions are path-identity matches and carry $k \geq 1$, which is exactly the sum of
the $k \geq 1$ counts above.

\begin{table}[ht]
\centering
\footnotesize
\begin{tabular}{lrr}
\toprule
Exposure step (MSE $\downarrow$) & Harmonizer & DoveNet \\
\midrule
Pooled, unmatched & $-106.98$ & $-70.90$ \\
Per-bin steps, sign & 8/8 negative & 8/8 positive \\
Per-bin step range & $-293.7$ to $-0.3$ & $+0.9$ to $+69.2$ \\
\bottomrule
\end{tabular}
\caption{Graded exposure dose-response for both baselines. Steps are $k \geq 1$ minus $k = 0$ inside
foreground-ratio bins, so a negative entry means the shared composite scores better. The bin-averaged
step and its detection floor are in main Table~2.}
\end{table}

\begin{table}[ht]
\centering
\footnotesize
\begin{tabular}{lrrrr}
\toprule
Method (MSE $\downarrow$) & ret. & excl. & offic. & $\Delta$ \\
 & 2{,}476 & 1{,}807 & n/a & excl.$-$ret. \\
\midrule
\multicolumn{5}{l}{\emph{HCOCO only, the population of the HCOCO-only matched analysis (main Table~2)}}\\
DoveNet & $98.96$ & $67.16$ & n/a & $-31.80$ \\
Harmonizer & $40.24$ & $24.92$ & n/a & $-15.32$ \\
Editor (ours) & $74.11$ & $47.00$ & n/a & $-27.12$ \\
\bottomrule
\end{tabular}
\caption{HCOCO-only score rows for the same three methods; the pooled version is main Table~1 and the matched, foreground-controlled version is main Table~2.}
\end{table}

\section{Fixed-model probe}

\noindent At five seeds the smallest attainable two-sided sign-test value is $p = 0.0625$, and the
in-region L1 gap is positive in all five seeds ($+1.954$ to $+2.938$).

\begin{table}[ht]
\centering
\footnotesize
\begin{tabular}{lrrr}
\toprule
 & Excluded & Retained & Contrast \\
\midrule
\multicolumn{4}{l}{\emph{Unmatched, full populations (1{,}823 / 5{,}581)}}\\
In-region L1 $\downarrow$ & $22.58$ & $25.01$ & 5/5 seeds \\
MSE $\downarrow$ & $48.82$ & $149.45$ & $3.06\times$ \\
fMSE $\downarrow$ & $871.77$ & $1104.82$ & $1.27\times$ \\
PSNR $\uparrow$ (dB) & $35.04$ & $31.91$ & $-3.13$ \\
Do-nothing L1 $\downarrow$ & $24.05$ & $29.64$ & $1.23\times$ \\
\midrule
\multicolumn{4}{l}{\emph{Within HCOCO: unmatched $\rightarrow$ foreground-matched [95\% CI]}}\\
MSE & $-27.12$ & \multicolumn{2}{r}{$-2.25$ $[-10.09, +6.20]$} \\
fMSE & $-52.45$ & \multicolumn{2}{r}{$-42.77$ $[-97.12, +15.56]$} \\
PSNR (dB) & $+1.30$ & \multicolumn{2}{r}{$+0.04$ $[-0.22, +0.30]$} \\
\bottomrule
\end{tabular}
\caption{The fixed-model probe. Top: five per-seed checkpoints scored on the two halves of the official list,
every contrast retained relative to excluded, the in-region gap positive in all five seeds. Bottom:
excluded-minus-retained differences inside HCOCO.}
\end{table}

\subsection*{S2.1 Resampling unit of the probe intervals}

The bottom panel of Table~S3 reports intervals from a cluster bootstrap whose resampling unit is the
target photograph, so every composite built on one photograph moves with it. The committed probe script
originally resampled composites within each foreground-ratio bin, which ignores that correlation and
returns narrower intervals. The script now implements the photograph unit as its default, and on the
recorded seed and configuration it reproduces the reported intervals to within bootstrap noise:

\begin{table}[H]
\centering
\small
\begin{tabular}{l r r r}
\toprule
Interval & Reported & Re-run, photograph clusters & Re-run, composite clusters \\
\midrule
MSE & $[-10.09, +6.20]$ & $[-9.92, +5.50]$ & $[-5.43, +1.25]$ \\
fMSE & $[-97.12, +15.56]$ & $[-96.43, +12.52]$ & $[-66.25, -19.01]$ \\
PSNR (dB) & $[-0.22, +0.30]$ & $[-0.22, +0.31]$ & $[-0.06, +0.15]$ \\
\bottomrule
\end{tabular}
\caption{The fixed-model probe intervals under the two resampling units, $2{,}000$ draws each, $8$
foreground-ratio bins, seed as recorded. The photograph unit reproduces the reported figures; the
composite unit does not, and it is the unit that matters here, because a test composite is correlated
with every training composite built on the same photograph.}
\label{tab:probe_units}
\end{table}

The residual differences between the reported and re-run photograph-clustered intervals are bootstrap
noise: $2{,}000$ draws over $3{,}417$ photograph clusters. The choice of unit is not cosmetic. Under
the composite unit the matched fMSE interval excludes zero, so a reader who took that unit would read a
detected difference where the photograph unit reads none. That is the reason the unit is stated here and
in the main text: the paper's argument is that composites sharing a photograph are not independent, and
the uncertainty calculation has to use the unit the audit keys on.
\section{Three-population scores for every checkpoint}

\noindent In every pooled bin $92.5$ to $100$ percent of the exposed composites are HCOCO, while the HCOCO share of the unexposed rises from $0.52$ in the first bin to $0.76$ in the second and then falls to $0.22$ in the eighth (largest-foreground) bin, and the two
largest-foreground bins carry $79$, $82$ and $80$ percent of the bin-averaged pooled step for the three
methods.

\begin{table}[htbp]
\centering
\small
\setlength{\tabcolsep}{4pt}
\begin{tabular}{lccccc}
\toprule
 & DoveNet & Harmonizer & PCT-Net & DucoNet & DiffHarmony \\
\midrule
MSE, official     & $122.75$ & $107.08$ & $78.56$ & $107.60$ & --- \\
fMSE, official    & $1132.68$ & $794.24$ & $608.26$ & $791.32$ & --- \\
MSE, retained     & $140.18$ & $133.42$ & $97.41$ & $135.04$ & $162.42$ \\
fMSE, retained    & $1114.71$ & $894.34$ & $682.00$ & $905.86$ & $523.90$ \\
MSE, excluded     & $69.40$ & $26.43$ & $20.83$ & $23.58$ & $196.67$ \\
fMSE, excluded    & $1187.67$ & $487.80$ & $382.50$ & $440.63$ & $512.95$ \\
\midrule
retained/excluded & $2.02$ & $5.05$ & $4.68$ & $5.73$ & $0.83$ \\
\bottomrule
\end{tabular}
\caption{Every checkpoint on the same three lists, one protocol, so the official column is the pooled number
and the other two are the populations composing it. DiffHarmony has no official row, its bundle not
having been re-extracted, and is reported descriptively.}
\label{tab:three_pop}
\end{table}

\begin{table}[htbp]
\centering
\small
\begin{tabular}{llrrr}
\toprule
Method & Population & MSE & fMSE & composites \\
\midrule
Editor (five seeds) & retained & $149.45$ & $1104.82$ & $27,905$ \\
Editor (five seeds) & excluded & $48.82$ & $871.77$ & $9,115$ \\
\bottomrule
\end{tabular}
\caption{The in-house typed editor, five per-seed checkpoints scored once each, for reference. It is a
control in this study and not a harmonization-quality claim; its bound carries slack and is below the
edit magnitude for two of the five seeds (Section~S15).}
\label{tab:editor_pop}
\end{table}

\section{Near-duplicate audit}

A 64-bit dHash is computed over every unique real background image: HCOCO 26{,}912 train
against 2{,}025 retained-test, HAdobe5k 3{,}897 against 432, HFlickr 4{,}397 against 436,
Hday2night 82 against 24. Test is queried against train at Hamming radii $\{0, 2, 4, 6, 8\}$,
and every Hamming-0 pair is confirmed by a $64 \times 64$ grayscale correlation.

\begin{table}[ht]
\centering
\scriptsize
\setlength{\tabcolsep}{5pt}
\begin{tabular}{lrr}
\toprule
Hamming radius & Pairs & Status \\
\midrule
0 & 4 & confirmed (1 HCOCO, 1 HAdobe5k, 2 HFlickr; three with $r \geq 0.9999$, one with $r = 0.9509$) \\
$\leq 2$ & 5 & candidate pairs \\
$\leq 4$ & 8 & candidate pairs \\
$\leq 6$ & 19 & candidate pairs \\
$\leq 8$ & 78 & candidate pairs \\
\bottomrule
\end{tabular}
\caption{Cross-split pairs beyond the registered exclusions. Wider radii retrieve unadjudicated candidates that
mix same-scene and re-encoded variants: neither the candidate counts nor the four confirmed pairs bound
total residual sharing from above. The confirmed floor is $0.14$ percent.}
\end{table}

\paragraph{Whole-matrix screen.} The hash radii above are bounded by the radii tested. A second,
independent pass compares every one of the 7{,}404 test composites with every one of the 65{,}742
training and validation composites at $32 \times 32$ grayscale correlation and reports cross-split
pairs above $0.99$ whose identity strings differ. It returns six candidates. One is byte-identical and
its composite is one of those removed by the decoded-pixel check. The other five sit in the retained list at correlation $0.99997$ or above; four are composites of three of the four confirmed pairs above, and only the fifth, a screen candidate rather than a confirmed pair, is additional. The screen is a correlation at a fixed
scale, so it is a floor on what a stricter comparison would find, not a proof of cleanliness.

\section{External comparison on ccHarmony}

\paragraph{Provenance of the numbers above.} The probe's excluded population reads $4.63$ against the
retained one, and the within-HCOCO interval is the one in the committed record: the committed probe
script reproduces its point estimates and not this interval.

\begin{table}[H]
\centering
\footnotesize
\begin{tabular}{lr}
\toprule
Descriptive quantity & Value \\
\midrule
Retained sign agreement & 5 of 6 \\
Official sign agreement & 6 of 6 \\
$\Delta_{\mathrm{agreement}}$ & $-1$ of 6 \\
LOMO without DucoNet: Delta sign & $0$ \\
LOMO without Harmonizer: Delta sign & $0$ \\
Kendall tau (retained, ccHarmony) & $0.6667$ \\
\quad 95\% interval & $[0.6667,1.0]$ \\
\bottomrule
\end{tabular}
\caption{External comparison (E2, the ccHarmony comparison of main Section~4.5): descriptive quantities within the frozen four-method roster. Signs use point gaps.
Two leave-one-method-out removals change the full-roster Delta sign to zero. This table
carries no efficacy verdict or external-validation claim.}
\end{table}

\section{Gate evidence and training configuration}

\noindent The second factor's capacity search overshoots the first by $2{,}592$ active parameters
($47{,}235$ against $44{,}643$), a width confound the paired ablation is free of by construction.

The editor is a compact CNN on $256 \times 256$ inputs, run at $(\gamma_P, \gamma_S) = (0.1, 0)$.
Five seeds $\{0,1,2,3,4\}$, halo 0, 15 epochs, $N = 1024$, batch 16, best-on-validation
checkpointing, strict cuDNN determinism in paper mode: the control arm measures
$\lvert \Delta \rvert = 1.849$ against the treatment at exactly $0.000$.

\begin{table}[ht]
\centering
\footnotesize
\begin{tabular}{lr}
\toprule
Quantity (held-out validation L1) & Value \\
\midrule
Typed channel vs capacity floor & $+0.358$ $[+0.071, +0.645]$ \\
\quad seeds positive & 5/5 \\
Paired $\gamma_P$ ablation (5 seeds) & $+0.694$ $[+0.510, +0.878]$ \\
\quad seed-to-seed sd & $0.148$ \\
\quad two-sided sign test & $p = 0.0625$ \\
Oracle signal, F1 $-$ F2 & $+0.994$ $[+0.856, +1.132]$ \\
Price of certification, F3 $-$ F2 & $+0.636$ $[+0.451, +0.822]$ \\
Certified share of oracle signal & $\approx 36$ \% \\
Physical certificates (range) & $0.049$ to $0.487$ \\
Semantic certificates & exactly $0$ \\
\bottomrule
\end{tabular}
\caption{Gate evidence and the three-factor decomposition, five seeds on held-out validation. The paired
ablation is the pre-registered primary estimator: identical shapes, initialization and batches, with
only the physical gate differing, so no selection enters it.}
\end{table}

Tightness: the held-out projected-gradient check (100 steps, $\varepsilon = 8/255$, 100 inputs
for each of the five seeds, 500 searches, zero soundness violations) gives per-seed ratios of
mean achieved change to mean bound of $0.070$, $0.077$, $0.094$, $0.073$, $0.072$ (mean $0.077$,
worst single input $0.28$). Table~S22 gives the physical-certificate bound at $\varepsilon = 8/255$ for all five
selected checkpoints.
It ranges from $1.85$ to $17.70$ levels.

\section{The third method in the standard protocol}

\paragraph{Provenance and record names.} The two public baselines are re-derived from pinned public
code and checkpoints in the re-derivation run \texttt{20260830-174720-baseline-rederivation}, over the
retained list \texttt{retained\_5581}.

\enlargethispage{1.2\baselineskip}
\begin{table}[H]
\centering
\footnotesize
\begin{tabular}{lrrrr}
\inputrows{t5_rows.tex}
\end{tabular}
\caption{Standard-protocol metrics on the retained list, every method on the identical list. Our rows are the
five-seed mean of the per-seed checkpoints and the two public baselines are re-derived from pinned code.}
\end{table}

\section{Cross-dataset construction scan}

\paragraph{Per-collection file counts.} ccHarmony file level: 4{,}260 composites present (3{,}080 and
1{,}180), 250 train and 100 test real files referenced, none missing, no real file or mask shared
across the split, 350 real files all referenced. DESOBA file level: zero missing files in four folders,
1{,}012 images equal to 839 $+$ 160 $+$ 13, and no CRC-identical image across the split. The DESOBA
shared stem is \texttt{web-shadow0248}, whose test \texttt{-1} and train \texttt{-2} read as
near-duplicate scene content at grayscale correlation $0.35$.

The construction scans use a name-level, CPU-only, read-only identity matcher. The scan script is released in the project repository as \texttt{dataset\_identity\_scan.py}, and the counts below are recorded values from it. Identity rule: list
entry without directory and extension, split on \texttt{\_}.
The last two tokens (variant letter
or mask id, and composite index) are dropped.
Identity is the first token, or the first two
tokens joined when the name has four or more tokens.
IHarmony4 identities are scoped by
sub-dataset.
DESOBA uses the file stem minus a trailing \texttt{-N} variant index.

\begin{table}[ht]
\centering
\scriptsize
\setlength{\tabcolsep}{4pt}
\begin{tabular}{lrrrr}
\toprule
Dataset (units) & Train / test units & Train / test identities & Shared & Test covered \\
\midrule
ccHarmony, published lists (composites) & 3{,}080 / 1{,}180 & 250 / 100 & 0 & 0 of 1{,}180 \\
iHarmony4 official lists (composites) & 65{,}742 / 7{,}404 & 35{,}288 / 4{,}317 & 1{,}399 & 1{,}822 of 7{,}404 (24.6\%) \\
iHarmony4 retained lists (composites) & 65{,}742 / 5{,}581 & 35{,}288 / 2{,}917 & 0 & 0 of 5{,}581 \\
ccHarmony test vs iHarmony4 official train & 1{,}180 test & 100 / 35{,}288 & 0 & 0 \\
DESOBA (source images) & 839 / 160 & 816 / 160 & 1 & 1 of 160 (0.6\%) \\
\bottomrule
\end{tabular}
\caption{Construction scan. The recorded iHarmony4 figure is 1{,}823 of 7{,}404, of which 1{,}822 are
path-identity composites and one is a decoded-pixel duplicate. The DESOBA shared stem reads as a
near-duplicate scene at grayscale correlation $0.35$, and ccHarmony shares no real file with the test
side.}
\end{table}

Blocked or not scanned: iHarmony8K (no public source found after three bounded searches).
DESOBAv2 (the only reachable archive is 12.69 GiB and was not fetched, and no split documentation
was found).
SRD, USR, WSID (no verified source).
ISTD+ (not a composite benchmark, not scanned).
HYouTube (not located, not scanned). The scan does not show whether identically named identities
are the same photograph, near-duplicate or re-photographed content across splits, foreground
overlap, leakage through real-image pools or pretraining data, or any effect on model scores.
The DESOBA pair counts (2{,}995 train and 624 test object-shadow pairs) are the README's and were
not re-derived. The ccHarmony cross-check against iHarmony4 has low power because the name
formats do not overlap.

\section{Provenance of the reported values}

\paragraph{Audit.} The four-checkpoint audit values (main Section~4.4) come from its retained
evaluation record. The MSE and PSNR sensitivity reads come from the retained MSE, PSNR and
metric-agreement records, kept with the sensitivity-grid record.

\paragraph{Retrain estimates and training check.} The difference-in-differences and training-check values of main Table~3 come from a replacement record generated after a fresh, detached scoring pass of the training check (called M8 in the repository records). A canonical aggregate record named in the
retrain plan does not exist.
The replacement record is used in its place, and we do not claim the
two would be interchangeable. An earlier M8 scoring directory is superseded and none of its values
is used. The fresh directory holds twelve row files and twelve sidecar files. A first verification
re-hashed the twelve scored weight files and checked the frozen list order and the checkpoint
bindings.
It did not regenerate the M8 estimator or the bootstrap and did not check the
authenticity of receipts written on other machines. A second, independent pass regenerated the
DiD and M8 point estimates, the 95 percent bootstrap endpoints, the bootstrap standard errors and
the MDEs from the replacement rows, and each equals its saved value exactly as a binary64 number.
Four qualifications remain. (i) The full-precision DoveNet M8 difference between the two scoring
passes is not zero.
The two-decimal displays agree. (ii) An alternate floating-point DiD
computation agrees with the saved value within $10^{-9}$ and is not literally equal. (iii) The MDE
file names the superseded evidence directory and records no digest of the script that wrote it,
so the identity of that script is not established.
Both DiD estimates and intervals in it match
the replacement record, and the current script reproduces the file. (iv) The evaluation row files
carry no individual byte hashes. These passes concern regeneration from the rows. They do not
resolve the execution departures of main Section~8.

\paragraph{ccHarmony comparison.} The values of main Section~4.5 and Table~S8 come from the
retained report, its manifest and an integrity-validation record. The report binds the manifest.
The manifest does not bind the report bytes in return, which the frozen protocol did not require.
One sentence describing P6 inside the saved report is out of date, and the agreement it reports
uses both the retained and the official inputs. No source record or experiment was changed for
this version of the paper.

\section{The identity-disjoint correction}

\begin{table}[H]
\centering
\footnotesize
\begin{tabular}{lr}
\toprule
Quantity & Value \\
\midrule
Target photographs in both lists (train $\cap$ test) & 1{,}399 \\
\quad of which HCOCO & 1{,}391 \\
\quad of which HFlickr & 8 \\
Official test composites & 7{,}404 \\
Excluded test composites & 1{,}823 \\
\quad path-identity exclusions & 1{,}822 \\
\quad decoded-pixel duplicate & 1 \\
Retained test composites & 5{,}581 \\
Share of official test excluded & 24.6 \% \\
Official train composites (untouched) & 65{,}742 \\
Staged counts, train / val / test & 52{,}411 / 13{,}331 / 5{,}581 \\
\bottomrule
\end{tabular}
\caption{The identity-disjoint correction, derived by deterministic file-list arithmetic over
the official release. Only the evaluation list changes; the train list is untouched, so
published checkpoints remain eligible for rescoring without retraining.}
\end{table}

\begin{table}[H]
\centering
\footnotesize
\begin{tabular}{lrrr}
\toprule
Method & Official & Retained & Change \\
\midrule
\multicolumn{4}{l}{\emph{MSE} $\downarrow$}\\
DoveNet & $122.75$ & $140.18$ & $+17.43$ \\
Harmonizer & $107.08$ & $133.42$ & $+26.34$ \\
\quad gap (DoveNet $-$ Harm.) & $+15.67$ & $+6.76$ & $2.3\times$ smaller \\
\midrule
\multicolumn{4}{l}{\emph{fMSE} $\downarrow$}\\
DoveNet & $1132.68$ & $1114.71$ & $-17.97$ \\
Harmonizer & $794.24$ & $894.34$ & $+100.10$ \\
\midrule
\multicolumn{4}{l}{\emph{PSNR} $\uparrow$ (dB)}\\
DoveNet & $32.74$ & $32.37$ & $-0.37$ \\
Harmonizer & $34.49$ & $33.43$ & $-1.06$ \\
\bottomrule
\end{tabular}
\caption{What changes if you adopt the list. The same checkpoints on the official and the retained list: MSE
and PSNR are worse for both methods by unequal amounts, the gap between them shrinks by $2.3$ times,
and on fMSE the two move in opposite directions.}
\end{table}

\paragraph{Worked example in the released lists.} One target photograph,
identifier \texttt{c344138}, carries three composites in the released lists: \texttt{c344138\_1885590\_1}
at line 18{,}902 of the official training list, and \texttt{c344138\_688383\_1} and
\texttt{c344138\_309886\_1} at lines 106 and 3{,}024 of the official test list. Both test composites are removed by the audit, the training composite stays, and the figure in the main text shows the first and the third of them sitting on the same photograph with a different recolouring.

\paragraph{List files and hashes.} Three files are read, as the dataset repository distributes them:
the official training list (65{,}742 rows, SHA-256 \texttt{d423036e...}), the official test list
(7{,}404 rows, SHA-256 \texttt{513ebcf6...}) and the retained test list this release publishes
(5{,}581 rows, SHA-256 \texttt{d2605e68...}). The full digests are in the project records.

\section{Soundness argument for the typed certificate}

\noindent\textbf{Setting and the two properties.} Masked local editors (Blended Diffusion, DiffEdit, ZONE and BrushNet, cited in the main
text) control which pixels are written, and compositing makes that exact. They do not state which outside pixels were
allowed to steer the pixels written inside the region. We separate two properties.
\emph{Footprint locality} is the write-set property: the composite returns the source exactly where
the effective write support is zero. It is not the same as the scoring region. Let
$R_{\mathrm{target}} = \{m > 0.5\}$ be the binary scoring region and let $\beta$ be the feathered
blend weight the composite applies. The composite zeroes the blend where $\beta$ falls below a
fixed feather threshold $\theta$, so the effective write support is
$R_{\mathrm{write}} = \{\beta \geq \theta\}$, which can extend outside $R_{\mathrm{target}}$
because the mask is feathered, so a pixel outside $R_{\mathrm{target}}$ with $\beta \geq \theta$
can receive part of the edit. Compositing returns the source exactly outside $R_{\mathrm{write}}$,
and only there. \emph{Sensitivity locality} is the causal property, and it is stated for the
target coordinates inside the fixed trusted window.

\begin{definition}[Sensitivity locality]
$E$ is $(C, \Omega)$-sensitivity-local at $(x, m, t)$ if, for the fixed scoring region
$R_{\mathrm{target}} \subseteq W$ and every perturbation $\delta$ supported on the exterior
$\Omega = \neg W$ of the trusted window,
$\lVert \pi_{R_{\mathrm{target}}} E(x + \delta, m, t) - \pi_{R_{\mathrm{target}}} E(x, m, t) \rVert \leq C \lVert \delta \rVert$.
\end{definition}

\begin{theorem}[Typed certificate]
Fix a mask $m$ and trusted window $W$ (the same for all inputs compared), let
$R_{\mathrm{target}} \subseteq W$ be the binary scoring region, and let the in-region output be
\[\pi_{R_{\mathrm{target}}} y = f_{\mathrm{local}}(x^W, m) + |\gamma_P| \tanh f_P(x^W, m, g_P(x)) + |\gamma_S| \tanh f_S(x^W, m, g_S(x)),\]
where $x^W$ agrees with $x$ on $W$ and is filled from $x_W$ alone, the physical sub-channel is fed
by the declassification map $P(x) = [\mu_{\mathrm{RGB}}, \sigma_{\mathrm{RGB}}]$ and the semantic
sub-channel by a pooled feature, and the gates $\gamma_P, \gamma_S$ are nonnegative (so $|\gamma_P| = \gamma_P$ and
$|\gamma_S| = \gamma_S$). Assume finite weights, an evaluation-mode network of $3 \times 3$ circular
convolutions with norm-preserving MaxMin activations on the stated square grid, and the
statistics/nonlinear interface of the reported checkpoints. Then
$L_{\mathrm{cross}}^P \leq \gamma_P \sqrt{2}\, L(\mathrm{enc}_P) L(\mathrm{syn}_P)$ and
$L_{\mathrm{cross}}^S \leq \gamma_S\, L(\mathrm{enc}_S) L(\mathrm{syn}_S)$.
In particular $\gamma_S = 0$ implies
$L_{\mathrm{cross}}^S \equiv 0$.
\end{theorem}

\noindent\textbf{Soundness outline.} The local path's out-of-window fill
depends only on $W$, so its gradient with respect to $x_\Omega$ is zero.
The mean pool's
$1/\sqrt{N}$ and the broadcast's $\sqrt{N}$ cancel exactly, and per-layer operator norms use the
FFT characterization of circular-convolution singular values (Sedghi et al., ICLR 2019), exact
at the grid they are computed on, with norm-preserving MaxMin activations (Anil et al., ICML
2019). The $\sqrt{2}$ is the declassification factor and does not depend on the pooling
size. The semantic certificate is zero by construction, a property of the computation
graph rather than a selected outcome. Soundness is verified three ways in the repository:
closed-form re-derivation, regression tests at odd resolutions with random $\Omega$-supported
perturbations, and a negative control that can fail. The full derivation, with norms,
units and hypotheses, is Section~S15.

Because the certificate upper-bounds how far admitted context can move the in-region edit, the bound
can be compared with a declared context budget: opening the gate wider to force a larger in-region
change raises $L_{\mathrm{cross}}$ and weakens the guarantee everywhere else. A budget-enforcing
edit-or-refuse decision rule that acts on this comparison is not evaluated here.
The only helper in the
code chooses a gate by a quality threshold and, if none passes, returns the richest-context gate,
which is not a fail-closed budget rule.

\section{Retrain design budgets}

The registered final-checkpoint budgets are 25 epochs and 1{,}580{,}650 steps for DoveNet, and 60
epochs and 474{,}180 steps for Harmonizer. Each arm has 63{,}226 training composites: a common 61{,}264-composite unexposed base plus a matched 1{,}962-composite arm set. On the test side, exposed comprises 1{,}874 composites.
Rest comprises 5{,}530
retained composites after removing 51 exposed overlaps.

\begin{figure}[H]
\centering
\includegraphics[width=0.6\textwidth]{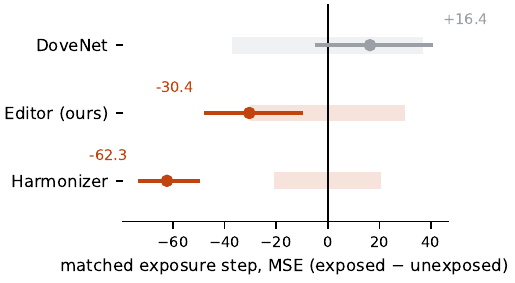}
\caption{The matched exposure step with its detection floor. Points are the bin-averaged step, bars the 95
percent interval and the pale band the detection floor. Pooled rows only: Harmonizer's and the editor's intervals lie below zero; DoveNet's includes zero.}
\end{figure}

\section{Retrain departures per arm}

This section records the departure classes and their claim consequences for the twelve retrain arms (Tables~S14 to S17). In the repository the full-list arm is named \texttt{leak\_in} and the reduced-list arm \texttt{leak\_out}; the tables below use the repository names. Per-arm run roots, manifest fields, weight hashes, environment records and
commit identifiers are kept in the project repository and are not reproduced here.

\begin{table}[H]
\centering
\scriptsize
\begin{tabular}{llp{4.1in}}
\toprule
Class & Name & Meaning \\
\midrule
X & Registered-item departure & Effective weight updates differ from the registered budget, whatever the manifest counter says \\
S & Stream continuity & Random stream or data order after a resume is not a continuation of the original, so the arm pair differs in more than exposure \\
E & Execution structure & Crash, pause, supervisor kill, separate invocations, uncollected attempts, co-tenancy \\
R & Runtime and environment & Machine, torch or python version, worker count, or recorded code revision differs between pairs \\
G & Provenance gap & A departure or tool is documented only in local-only files, or a hash or check output is missing \\
P & Post-tag collection & M8 scoring did not exist at the tag \\
\bottomrule
\end{tabular}
\caption{Departure classes used in Tables~S15 to S17.}
\end{table}

The values behind the departure classes come from the manifests and metrics sidecars of the
collected arms.

Constant across the twelve arms, read from every manifest: the evaluation and training list hashes
equal the registered ones, the confirmatory epoch equals the final epoch, and the scored row counts
are 5{,}581 and 1{,}874. The \texttt{is\_pre\_registered} flag is a driver-side flag; it is not
evidence that the departures above did not occur.

\begin{table}[H]
\centering
\scriptsize
\begin{tabular}{lp{5.4in}}
\toprule
Class & What the saved numbers may and may not be used for \\
\midrule
X & The affected arms' fMSE rows may be reported as measured values at the stated checkpoint and lists, and may not be described as the product of the registered step budget or of a confirmatory arm. \\
S & A \texttt{leak\_in} minus \texttt{leak\_out} contrast involving a resumed arm may be reported as a difference between two training histories that includes exposure, and may not be described as an exposure-only difference or as a clean paired-seed contrast. \\
E & The manifests may be cited for what each arm recorded, and may not be cited as showing one uninterrupted invocation per pair or a driver-verified equal-steps check. \\
R & Pair-level numbers may be compared within a pair, and may not be pooled or compared across seeds as if run in one environment. \\
G & Facts that rest on local-only files may be stated as locally recorded, and may not be offered as shipped or independently reproducible evidence. \\
P & M8 may be used to say the trained checkpoints differ on composites that sit in one arm's training set and not the other's, and may not be described as a pre-registered check passed at the tag. \\
\bottomrule
\end{tabular}
\caption{Claim consequences per departure class, one sentence each.}
\end{table}

\begin{table}[H]
\centering
\scriptsize
\begin{tabular}{lp{4.7in}}
\toprule
Arm & Consequence \\
\midrule
dovenet leak\_in s0 & The saved rows are a measured value with no recorded interruption; paired with leak\_out s0 they support no exposure-only reading. \\
dovenet leak\_out s0 & The saved rows come from a reseeded run that applied about 2.1 percent more updates than registered; they are descriptive and are not a confirmatory arm. \\
dovenet leak\_in s1 & The saved rows come from a crash-and-resume run that applied about 1.0 percent more updates than registered and whose manifest was edited by hand to resume; they are descriptive only. \\
dovenet leak\_out s1 & The saved rows have no resume record but trained concurrently with its partner arm and a duplicate run; they support no claim of an isolated or uninterrupted invocation. \\
dovenet leak\_in s2 & Same as leak\_in s0. \\
dovenet leak\_out s2 & Same as leak\_out s0. \\
harmonizer leak\_in s0 & The saved rows are a measured value with no recorded interruption, trained in a different invocation from its partner. \\
harmonizer leak\_out s0 & The saved rows come from a reseeded resume from checkpoint 20 after a discarded first pass of about six epochs, preceded by an abandoned attempt; they are descriptive only. \\
harmonizer leak\_in s1 & The saved rows are an uninterrupted single-process run on laptop A, which does not carry over to its partner. \\
harmonizer leak\_out s1 & The saved rows come from three resumes, two without RNG restoration, with ten epochs of first-pass work discarded; they support no uninterrupted-RNG or exposure-only claim. \\
harmonizer leak\_in s2 & The saved rows are an uninterrupted single-process run on laptop B, which does not carry over to its partner. \\
harmonizer leak\_out s2 & The saved rows come from a resume at checkpoint 10 with recorded RNG restoration; bitwise continuity and an exposure-only contrast are not established. \\
\bottomrule
\end{tabular}
\caption{Claim consequences per arm, one sentence each.}
\end{table}

\begin{table}[H]
\centering
\scriptsize
\begin{tabular}{lp{5.3in}}
\toprule
Method & Consequence \\
\midrule
DoveNet & All three pairs hold a class X arm, so the DoveNet DiD and its MDE are reported with neither difference nor equivalence established, and the pre-registered prediction ``equivalent'' is not supported. \\
Harmonizer & All three pairs hold a resumed or reseeded \texttt{leak\_out} arm, so the Harmonizer DiD and its MDE are reported with neither difference nor equivalence established, and the pre-registered prediction ``different, negative'' is not supported. \\
Both & No method verdict, no causal null, no equivalence, no confirmed prediction; the word ``confirmatory'' is not used as a claim about either method. \\
\bottomrule
\end{tabular}
\caption{Claim consequences at the method level.}
\end{table}

\section{Post-hoc sub-dataset check of the exposure step}

\paragraph{Numbers behind the random-removal reference.} Random removal of 1{,}807 HCOCO composites
plus the 16 other excluded ones gives a remaining-set MSE gap of $2.07 / 3.69 / 5.14$ and an
official-to-remaining ratio of $3.05 / 4.25 / 7.57$ at the 2.5th, 50th and 97.5th percentile, against
the observed $2.32$.

{\emergencystretch=3em
Label: post-hoc re-analysis of committed per-composite records, not pre-registered, CPU only, no
model run. The retained record and tables use the estimator behind main Table~2, reused unchanged:
eight
quantile foreground-ratio bins computed on the records passed in, $k \geq 1$ minus $k = 0$ inside
each bin, equal-weight mean over bins holding both arms, identity-cluster bootstrap, MDE
$= 2.8 \times$ bootstrap SE. Here the new strata use 2{,}000 draws.
Main Table~2 pooled rows use
400 draws and were reproduced exactly at 400 draws (rng seed 11). A stratum is called estimable
only with at least 30 unexposed and 30 exposed composites. PCT-Net, DucoNet and DiffHarmony are
the registered-audit (E1, main Section~4.4) per-composite rows.
DiffHarmony is outside the frozen four-method family and descriptive.
No interval below is adjusted for multiplicity.\par}

\begin{table}[H]
\centering
\scriptsize
\resizebox{\linewidth}{!}{\begin{tabular}{llrrrrr}
\toprule
Method & Metric & Pooled step [95\% CI] & HCOCO-only step [95\% CI] & HCOCO MDE & Pooled minus HCOCO [95\% CI] & Share \\
\midrule
DoveNet & MSE & $+16.36$ $[-6.4, +40.1]$ & $+2.08$ $[-8.5, +13.4]$ & $15.7$ & $+14.27$ $[-2.4, +32.7]$ & 87\% \\
Harmonizer & MSE & $-62.31$ $[-74.9, -49.7]$ & $-2.86$ $[-7.0, +1.3]$ & $5.9$ & $-59.45$ $[-71.6, -47.6]$ & 95\% \\
Editor (ours) & MSE & $-30.40$ $[-47.3, -10.7]$ & $-2.32$ $[-10.0, +5.8]$ & $11.0$ & $-28.08$ $[-42.2, -11.4]$ & 92\% \\
PCT-Net & MSE & $-42.49$ $[-52.0, -32.5]$ & $-1.40$ $[-4.8, +2.1]$ & $4.9$ & $-41.09$ $[-49.5, -32.4]$ & 97\% \\
DucoNet & MSE & $-69.35$ $[-82.4, -56.6]$ & $-1.72$ $[-5.4, +2.1]$ & $5.4$ & $-67.63$ $[-79.8, -56.0]$ & 98\% \\
DiffHarmony & MSE & $+50.23$ $[+41.0, +59.3]$ & $+26.76$ $[+19.5, +34.3]$ & $10.7$ & $+23.47$ $[+16.5, +30.4]$ & 47\% \\
\midrule
DoveNet & fMSE & $+134.30$ $[+47.4, +223.1]$ & $-40.50$ $[-120.2, +40.2]$ & $115.0$ & $+174.80$ $[+107.2, +240.2]$ & 130\% \\
Harmonizer & fMSE & $-368.30$ $[-420.3, -314.2]$ & $-50.33$ $[-82.5, -16.2]$ & $46.2$ & $-317.98$ $[-368.1, -270.8]$ & 86\% \\
Editor (ours) & fMSE & $-188.46$ $[-251.5, -116.5]$ & $-43.78$ $[-98.4, +10.4]$ & $78.4$ & $-144.67$ $[-200.4, -90.2]$ & 77\% \\
PCT-Net & fMSE & $-272.49$ $[-313.2, -227.1]$ & $-44.53$ $[-73.9, -15.9]$ & $41.4$ & $-227.96$ $[-264.0, -190.4]$ & 84\% \\
DucoNet & fMSE & $-427.43$ $[-485.4, -369.5]$ & $-35.65$ $[-65.7, -3.8]$ & $44.0$ & $-391.78$ $[-445.8, -340.1]$ & 92\% \\
DiffHarmony & fMSE & $-44.58$ $[-84.0, -7.5]$ & $-78.15$ $[-124.2, -33.4]$ & $63.3$ & $+33.57$ $[+2.4, +63.4]$ & $-75$\% \\
\bottomrule
\end{tabular}}
\caption{Pooled and HCOCO-only matched steps. Share is the pooled-minus-HCOCO difference over the pooled step.
Inside HCOCO the steps of the three methods are not detected under this estimator and uncertainty
calculation, so they are not read as an association with exposure.}
\end{table}

\begin{table}[H]
\centering
\scriptsize
\begin{tabular}{lrrrr}
\toprule
Method (MSE) & Pooled retained / excluded [95\% CI] & HCOCO retained / excluded [95\% CI] \\
\midrule
DoveNet & $2.02$ $[1.75, 2.34]$ & $1.47$ $[1.29, 1.69]$ \\
Harmonizer & $5.05$ $[4.27, 5.88]$ & $1.61$ $[1.41, 1.84]$ \\
Editor (ours) & $3.06$ $[2.63, 3.55]$ & $1.58$ $[1.38, 1.80]$ \\
PCT-Net & $4.68$ $[3.95, 5.52]$ & $1.56$ $[1.36, 1.78]$ \\
DucoNet & $5.73$ $[4.82, 6.75]$ & $1.56$ $[1.37, 1.77]$ \\
DiffHarmony & $0.83$ $[0.79, 0.87]$ & $0.90$ $[0.86, 0.93]$ \\
\bottomrule
\end{tabular}
\caption{Whole-image MSE ratios of the retained to the excluded list, pooled and inside HCOCO
(identity-cluster bootstrap, 2{,}000 draws). The HCOCO ratios are not foreground-matched.
DiffHarmony scores worse on the excluded list in both populations.}
\end{table}

\begin{table}[H]
\centering
\scriptsize
\resizebox{\linewidth}{!}{\scriptsize
\begin{tabular}{lcccc}
\toprule
 & Pooled MSE & HCOCO MSE & Pooled fMSE & HCOCO fMSE \\
\midrule
Retained gap & $6.76$ $[-5.20, +18.74]$ & $58.73$ $[+52.38, +65.23]$ & $220.38$ $[+166.64, +276.28]$ & $710.70$ $[+667.26, +755.52]$ \\
Excluded gap & $42.97$ $[+37.43, +49.26]$ & $42.25$ $[+37.20, +48.11]$ & $699.88$ $[+648.42, +752.37]$ & $700.02$ $[+648.47, +753.15]$ \\
Official gap & $15.67$ $[+6.44, +24.82]$ & $51.77$ $[+47.55, +56.05]$ & $338.44$ $[+294.60, +384.20]$ & $706.19$ $[+669.98, +738.56]$ \\
Official / retained & $2.32$ $[-13.42, +15.16]$ & $0.88$ $[+0.84, +0.94]$ & $1.54$ $[+1.37, +1.79]$ & $0.99$ $[+0.96, +1.03]$ \\
Excluded $-$ retained & $+36.21$ $[+22.87, +50.25]$ & $-16.48$ $[-24.56, -8.32]$ & $+479.50$ $[+403.61, +557.22]$ & $-10.68$ $[-77.97, +56.26]$ \\
\bottomrule
\end{tabular}}
\caption{DoveNet-minus-Harmonizer gaps on the same composites. The pooled fMSE interaction is 479.50. Random
removal of 1{,}807 HCOCO composites plus the other 16 gives an official-to-remaining ratio of 3.05 to
7.57 against the observed 2.32, a reference rather than a matched placebo.}
\end{table}

\begin{table}[H]
\centering
\scriptsize
\begin{tabular}{lrrrrr}
\toprule
Sub-dataset & Composites & Retained & Excluded & $k = 0$ & $k \geq 1$ \\
\midrule
HCOCO & 4{,}283 & 2{,}476 & 1{,}807 & 2{,}477 & 1{,}806 \\
HAdobe5k & 2{,}160 & 2{,}160 & 0 & 2{,}160 & 0 \\
HFlickr & 828 & 812 & 16 & 812 & 16 \\
Hday2night & 133 & 133 & 0 & 133 & 0 \\
\midrule
\multicolumn{6}{l}{Pooled foreground bins (edges from the 7{,}404 composites): HCOCO share of the unexposed / exposed composites} \\
Bin 0 to 3 & \multicolumn{5}{l}{$0.521 / 0.994$, $0.756 / 1.000$, $0.607 / 1.000$, $0.472 / 0.985$} \\
Bin 4 to 7 & \multicolumn{5}{l}{$0.411 / 1.000$, $0.420 / 0.989$, $0.312 / 0.964$, $0.220 / 0.925$} \\
\bottomrule
\end{tabular}
\caption{Exposure by sub-dataset, one copy of each composite. HAdobe5k and Hday2night contain no exposed
composite and HFlickr's 16 are below the 30-composite rule. Below, the largest-foreground bins carry
most of the pooled step.}
\end{table}

\paragraph{Scope.} Nothing here is a causal statement. Exposed and unexposed composites inside
HCOCO are not randomized. Among sub-datasets, only HCOCO passes the 30-composite reliability rule. The bins are the
estimator's eight quantile bins with no other matching, and the 72 stratum records, of which 48 (all populations with at least one exposed composite) enter the H5 family and 24 (pooled and HCOCO only) pass the 30-unexposed/30-exposed rule above, carry no family-wise control at the per-bin level. The tables of this section fit no dose-response
beyond $k = 1$ and pool the five seeds of the typed editor. Both were computed post-hoc by the statistics script of Section~S16. No slope of error on $k$
among exposed composites has an interval excluding zero for any of the three main Table~2
methods, in MSE or fMSE, under the specifications tried (linear and $\log_2 k$, equal-weight and
fixed-effects, pooled and HCOCO bins, and bins 0 to 6 only).
The detection floors are $8.6$ to
$56.4$ MSE per extra copy, so this is an absence of detection and not a flat dose-response.
Per seed (2{,}000 draws each), the pooled MSE step of the typed editor is $-27.32$ to $-33.35$
(five of five negative, every interval below zero, seed mean $-30.40$) and its HCOCO-only MSE
step is $-1.47$ to $-2.89$ (every interval includes zero). The seeds repeat the pooled picture
and are not independent composition estimates.

\begingroup\emergencystretch=4em
\section{Proof of the typed cross-region certificate}

This section proves the inequality of the theorem stated in Section~S11 for the code path that produced the reported
checkpoints and states the hypotheses it needs. Every step follows the implementation's editor,
Lipschitz and evaluation components. No reported
value is changed. The statements concern the real-valued network and the float compositing path.
The integer images of the audit are treated in S15.4.

\paragraph{S15.1 Setting}
\emph{Scale.} The network works on floats in $[0,1]$. The editor divides the uint8 image by $255$,
so a budget of $\varepsilon$ levels (0 to 255) is $\varepsilon_{01} = \varepsilon / 255$ and a
network-scale quantity $q$ corresponds to $255\,q$ levels. The image is square, $n \times n$ with
$n = 256$ (the audit passes the image height as the FFT grid), and $N = n^2$ is the pixel count.
This $N$ is not the $N = 1024$ samples per split of S6.

\emph{Norms.} $\lVert \cdot \rVert_2$ is the Euclidean norm over all $3N$ entries (channels and
positions), $\lVert \cdot \rVert_\infty$ the largest absolute entry. Every constant $L$ below is
an operator norm from $\lVert \cdot \rVert_2$ to $\lVert \cdot \rVert_2$ and is dimensionless.

\emph{Sets.} $R_{\mathrm{target}} = \{\text{mask} > 0.5\}$ with $n_R \geq 1$ pixels is the binary
scoring region. The composite feathers the mask with blend weight $\beta \in [0,1]$ and hard
preserves (blend set to zero) where the feathered weight falls below \path{outside_threshold}
$\theta$, so the effective write support is $R_{\mathrm{write}} = \{\beta \geq \theta\}$, which can
extend outside $R_{\mathrm{target}}$.
$W = R_{\mathrm{target}}$ dilated by the halo (\path{dilate_window}, halo $0$ gives
$W = R_{\mathrm{target}}$, the setting of every reported checkpoint). The statement covers
target coordinates in the fixed trusted window $W$, and the exact footprint equality below holds
only outside $R_{\mathrm{write}}$. $\Omega = \neg W$ with $n_\Omega \geq 1$ pixels. The mask $m$,
the halo and the feather settings are the same for all inputs compared.
The composite defaults to
feather radius $3$ and outside threshold $0.01$ (\path{cpu_editor.py}), and the wrapper settings of
the reported checkpoints were not independently replayed here.

\emph{Perturbation class.} For an image $x$, the admitted images are $x' \in [0,1]^{3 \times n
\times n}$ with $x'_W = x_W$, so $d = x' - x$ is supported on $\Omega$. This contains every additive
perturbation followed by clipping to $[0,1]$ (clipping cannot enlarge $\lvert d \rvert$). Changes
inside $W$ are outside the claim by construction.

\emph{The map.} Let $x^W$ be $x$ with every $\Omega$ pixel set to the per-channel mean of $x$ over
$W$ (the default fill of \path{GLWEditor.edit}).
It is a function of $x_W$ alone. Write
$P(x) = [\mu_c(x), \sigma_c(x)]_{c \in \mathrm{RGB}} \in \mathbb{R}^6$ with
$\sigma_c = \sqrt{\mathrm{var}_c + 10^{-12}}$, and $\mathrm{bc}$ for broadcast of a vector over the
$n \times n$ grid. The residual is, as in \path{TypedGLWNet.forward},
\begin{align*}
F(x) &= f_{\mathrm{loc}}([x^W, m]) + \gamma_P \tanh S_P([x^W, m, g_P(x)]) + \gamma_S \tanh S_S([x^W, m, g_S(x)]),\\
g_P(x) &= \mathrm{bc}\,\tanh E_P(P(x)), \qquad
g_S(x) = \mathrm{bc}\,\tanh \big(\tfrac{1}{N} \textstyle\sum_{\mathrm{pos}} E_S(x)\big).
\end{align*}
$E_P$ is a stack of $1 \times 1$ convolutions with MaxMin on the $6 \times 1 \times 1$ map.
$f_{\mathrm{loc}}$, $S_P$, $S_S$, $E_S$ are stacks of $3 \times 3$ circular convolutions with
MaxMin. The edit is $\mathrm{clamp}(x + F(x), 0, 1)$ followed by the feathered composite.

\begin{sloppypar}
\emph{Hypotheses.} (H1) The configuration is \texttt{physical\_encoder = statistics},
\texttt{physical\_interface = nonlinear} (the architecture stored in the reported checkpoints).
The other encoder and interface kinds in the code are not covered. (H2) The network is in
evaluation mode. It contains only convolutions, MaxMin and $\tanh$, no normalization or dropout,
\path{model_certified_lipschitz} raises an error for any other parameterized module. (H3) All
weights are finite. (H4) The FFT constants are computed at the grid $n$ of the input.
They are
not claimed for another size or for non-square images.
\end{sloppypar}

\paragraph{S15.2 Statement and proof}
\noindent\textbf{Claim 1 (the declassified statistic).}
$\lVert P(x') - P(x) \rVert_2 \leq \lVert d \rVert_2 / \sqrt{N}$ for all $x, x'$. The code uses the
weaker constant $\sqrt{2/N}$ (\path{declassified_statistics}), which is also valid.

\emph{Proof.} Fix a channel.
Let $a, a' \in \mathbb{R}^N$ be its values, $\Pi = \mathbf{1}\mathbf{1}^\top / N$,
$C = I - \Pi$. Then $\mu = \mathbf{1}^\top a / N$ gives $\lvert \mu' - \mu \rvert = \lVert \Pi (a' - a) \rVert_2 / \sqrt{N}$.
Also $\sigma = \lVert (C a / \sqrt{N},\, 10^{-6}) \rVert_2$ is a norm of an affine image of $a$, so the
reverse triangle inequality gives $\lvert \sigma' - \sigma \rvert \leq \lVert C (a' - a) \rVert_2 / \sqrt{N}$.
Summing squares over channels and using $\lVert \Pi v \rVert^2 + \lVert C v \rVert^2 = \lVert v \rVert^2$
gives $\lVert P(x') - P(x) \rVert_2^2 \leq \lVert d \rVert_2^2 / N$. $\square$

\noindent\textbf{Claim 2 (layer norms).} (a) $\tanh$ and MaxMin are $1$-Lipschitz for
$\lVert \cdot \rVert_2$ (MaxMin sorts disjoint channel pairs, and sorting does not expand the
Euclidean distance). (b) A stack of $1 \times 1$ convolutions on a $1 \times 1$ map has Lipschitz
constant at most the product of the spectral norms of its weight matrices
(\path{model_lipschitz_bound}).
Call this $\mathrm{Lip}(E_P)$. (c) A $3 \times 3$ convolution with
circular padding on the $n \times n$ grid is a circular convolution.
Its operator norm is the
largest singular value over the $n^2$ per-frequency transfer matrices (Sedghi et al.,
\path{conv_operator_norm_fft}), exact at that grid. The product over the layers of a stack bounds
the stack (sub-multiplicativity).
Call these $\mathrm{Lip}(S_P)$, $\mathrm{Lip}(S_S)$,
$\mathrm{Lip}(E_S)$. Biases cancel in differences.

\noindent\textbf{Theorem (network level).} Under H1 to H4, for every admitted $x'$,
\[
\begin{gathered}
\lVert F(x') - F(x) \rVert_2 \;\leq\; \big(L^P + L^S\big)\,\lVert d \rVert_2, \\
L^P = \sqrt{2}\,\lvert\gamma_P\rvert\,\mathrm{Lip}(E_P)\,\mathrm{Lip}(S_P), \quad
L^S = \lvert\gamma_S\rvert\,\mathrm{Lip}(E_S)\,\mathrm{Lip}(S_S),
\end{gathered}
\]
and, entrywise, $\lvert F(x') - F(x) \rvert \leq 2(\lvert\gamma_P\rvert + \lvert\gamma_S\rvert)$. These
are the quantities returned by \path{cross_region_lipschitz_physical}, \path{cross_region_lipschitz_semantic}
and their sum \path{cross_region_lipschitz}, with \path{input_size} $= n$. The bound holds on all
positions, hence on $R$.

\emph{Proof.} (1) \emph{Local term.} $x'^W = x^W$ because $x'_W = x_W$ and the fill is a function of
$x_W$, $m$ is fixed. So $f_{\mathrm{loc}}$ contributes $0$.
(2) \emph{Physical term.} The mean over a $1 \times 1$ map in \path{_broadcast_context} is the
identity. By Claim 2, $\lVert \tanh E_P(P(x')) - \tanh E_P(P(x)) \rVert_2 \leq \mathrm{Lip}(E_P)
\lVert P(x') - P(x) \rVert_2$. Broadcast over $N$ positions multiplies the Euclidean norm by
$\sqrt{N}$, so with Claim 1 in the code's form,
$\lVert g_P(x') - g_P(x) \rVert_2 \leq \sqrt{N}\,\mathrm{Lip}(E_P)\sqrt{2/N}\,\lVert d \rVert_2 =
\sqrt{2}\,\mathrm{Lip}(E_P)\lVert d \rVert_2$. The factor $\sqrt{N}$ from broadcasting and the factor
$1/\sqrt{N}$ from $P$ cancel at every resolution. The input of $S_P$ changes only in the $g_P$
block, and $\mathrm{Lip}(S_P)$ is an operator norm over the whole input, so it bounds the change
from that block alone. $\tanh$ is $1$-Lipschitz.
Multiply by $\lvert\gamma_P\rvert$.
(3) \emph{Semantic term.} For each of the $16$ context channels, the mean over positions changes by
at most $\lVert \Delta E_S \rVert_2 / \sqrt{N}$ (Cauchy-Schwarz), so the pooled vector changes by at
most $\mathrm{Lip}(E_S)\lVert d \rVert_2 / \sqrt{N}$. After $\tanh$ and broadcast ($\times \sqrt{N}$)
the factor $\sqrt{N}$ cancels again, giving $\lVert g_S(x') - g_S(x) \rVert_2 \leq \mathrm{Lip}(E_S)
\lVert d \rVert_2$.
Then $S_S$, $\tanh$ and $\lvert\gamma_S\rvert$ as in step 2.
(4) Add the two terms (triangle inequality). The entrywise bound is the range of $\tanh$:
each gated term moves an entry by less than $2\lvert\gamma\rvert$. $\square$

\noindent\textbf{Corollary 1 (edit, in-region).} Let $e(x) = \mathrm{clamp}(x + F(x), 0, 1)$ and let
$\beta \in [0,1]$ be the blend weight of the feathered composite. For $p \in R_{\mathrm{target}}$
the composite is $\beta(p)\,e(p) + (1 - \beta(p))\,x(p)$, and $x'(p) = x(p)$ because
$R_{\mathrm{target}} \subseteq W$. Clamp is $1$-Lipschitz and $\beta \leq 1$, so
$\lVert (\text{out}(x') - \text{out}(x))_{R_{\mathrm{target}}} \rVert_2 \leq (L^P + L^S) \lVert d
\rVert_2$ before integer rounding. Outside $R_{\mathrm{write}} = \{\beta \geq \theta\}$ the
composite returns $x$ exactly, which is the footprint statement.
A pixel with $\beta \geq \theta$
that lies outside $R_{\mathrm{target}}$ can still receive part of the edit, and under an admitted
perturbation the value of an $\Omega$ pixel can change, so $R_{\mathrm{write}}$ is not the
theorem's target set.

\noindent\textbf{Corollary 2 (budget in mean absolute units).} If $\lVert d \rVert_\infty \leq
\varepsilon_{01}$ then $\lVert d \rVert_2 \leq \varepsilon_{01}\sqrt{3 n_\Omega}$. The mean absolute
in-region change per pixel and channel is at most the root mean square, so with the per-channel
tanh range as a second bound,
\[
\bar{\Delta}_R \;\leq\; b^P + b^S, \qquad b^{\bullet} = \min\!\Big(2\lvert\gamma_\bullet\rvert,\;
L^{\bullet}\,\varepsilon_{01}\sqrt{n_\Omega / n_R}\Big).
\]
This is \path{certified_influence_typed}, $255\,(b^P + b^S)$ is the audit column \texttt{cert255} and
the quantity \texttt{mean\_bound} of the tightness check (times $255$). The geometric factor
$\sqrt{n_\Omega / n_R}$ converts the Euclidean budget of the theorem to the per-pixel mean the
audit measures. It is large when the region is small, which is why the table below is far from
$L \varepsilon_{01}$.

\paragraph{S15.3 Zero semantic term and physical dependence}
(i) With $\gamma_S = 0$ exactly, the semantic output term is multiplied by $0$ (finite $\tanh$
values, H3), so $L^S = 0$ and no path leads from $\Omega$ through the semantic modules. This is a
statement about the computation graph: the unrestricted learned-context branch is disabled and does
not influence the output, and nothing about that branch is certified. It does not imply that
exterior semantic content is absent. The channel labels are implementation names.
The remaining
exterior dependence is six global color moments, and those moments can correlate with semantic
content, so this is not a disentanglement claim. In the five reported checkpoints
\texttt{L\_cert\_semantic} is $0.0$.
(ii) The physical channel reads $\Omega$ only through $P(x)$. If $P(x') = P(x)$ the physical term is
identical: any permutation of $\Omega$ pixel values leaves every per-channel mean and standard
deviation unchanged. The regression test \texttt{test\_declassified\_channel\_is\_invariant\_to\_spatial\_semantic\_rearrangement}
checks the statistic to $10^{-7}$ and the edited region exactly.
In float arithmetic equality holds only up to summation order. If
$P(x') \neq P(x)$ the output can move, by at most $L^P \lVert d \rVert_2$: a uniform shift of
$\Omega$ changes the channel means and therefore the edit. So at $\gamma_S = 0$ the editor is not
independent of $\Omega$.
It depends on six numbers, with the stated bound. (iii) At
$\gamma_P = \gamma_S = 0$ the output is independent of $\Omega$ (test \texttt{test\_typed\_local\_write\_ignores\_omega\_when\_global\_gates\_zero}).

\paragraph{S15.4 Code checks and scope}
\emph{Pool and broadcast.} The mean-pool/broadcast cancellation of step 3 applies to the semantic
channel. The physical channel has a different cancellation: broadcast $\sqrt{N}$ against the
statistic's $1/\sqrt{N}$, leaving $\sqrt{2}$ with the code's constant. \emph{No downsample factor.}
The current source contains no \texttt{avg\_pool2d}. The pooled low-pass encoder whose factor $1/k$ cancelled
$\sqrt{N/N_p} = k$ was retired, \texttt{phys\_downsample} is kept in the constructor and checkpoint
metadata only and is not read by the forward pass. The certificate therefore does not depend on a
downsample factor. \emph{Constant.} By Claim 1 the factor $\sqrt{2}$ is valid but not tight,
$1$ would also be valid. It is not applied here and no reported value uses it.
\emph{Slack in $\mathrm{Lip}(S)$.} It is the norm over the whole input, an upper bound for the $g$
block. \emph{Resolution.} The FFT norm is exact only at the grid it is computed on, so the constants
are stated at $n = 256$ only, and it is not monotone in the grid size in general: non-nested
frequency grids can miss different maxima, so a norm at one size is not a bound at another. A scalar
kernel whose only nonzero row is $[1, -1, 0]$ has circular operator norm $2$ at $n = 4$ but
$2\cos(\pi/10) < 2$ at $n = 5$. Certification is therefore required at the resolution actually
evaluated. \emph{Integer rounding.} The editor truncates to uint8 twice (the edited image and the
composite) and the audit truncates the perturbed input. Truncation can move a value by up to one
level while the real-valued change is arbitrarily small, so audit columns measured on uint8 outputs
can exceed $255\,(b^P + b^S)$ by about one level per truncation.
The corresponding regression test
allows a margin of one level. The projected-gradient tightness check measures the float network
residual and has no such margin. \emph{Arithmetic.} The proof is for real arithmetic.
The constants
are float32 FFT and singular-value computations with no outward rounding, so the computed value is an
estimate and not an interval-certified machine bound, and we state no numeric rounding allowance
here. A CPU recomputation from the saved checkpoints reproduced the saved \texttt{mean\_bound} values
to a relative difference below $10^{-5}$. \emph{Role of PGD.} The proof carries soundness. The
search (S5) is a consistency check of the code against the proof, and with about $13\times$ slack it
cannot detect a bound that is wrong by a small factor.
\emph{Not covered.} Halo values other than $0$ were not part of the reported runs.
The proof does not
use the halo value. Nothing is claimed for editors outside H1 or for perturbations that touch $W$.

\paragraph{S15.5 Units for the reported checkpoints}
Table~S22 applies Corollary 2 to the five selected checkpoints. $L^P$ is the selected point's
\texttt{L\_cert\_physical} (the range of Table~S9). $8L^P$ is $255\,L^P\varepsilon_{01}$ for
$\varepsilon = 8$ levels, the value before the geometric factor. The next column is the saved
\texttt{mean\_bound} of the tightness check times $255$, a mean over $100$ inputs of the retained
list.
The last column is its ratio to $8L^P$. Where the tanh cap $2\lvert\gamma_P\rvert = 0.2$
($51$ levels) binds for some input, the mean falls below the product and the ratio falls below the common value
of the two uncapped seeds ($4.73$, the mean of $\sqrt{n_\Omega / n_R}$ over those $100$ inputs). The edit
magnitude is the validation mean absolute in-region edit of the selected point.
It comes from a
different input set than the bound.

\begin{table}[H]
\centering
\footnotesize
\begin{tabular}{rrrrrr}
\toprule
Seed & $L^P$ & $8L^P$ & Mean bound (levels) & Bound $/\, 8L^P$ & Edit magnitude (levels) \\
\midrule
0 & $0.4170$ & $3.34$ & $15.37$ & $4.61$ & $9.43$ \\
1 & $0.0758$ & $0.61$ & $2.87$ & $4.73$ & $8.98$ \\
2 & $0.0489$ & $0.39$ & $1.85$ & $4.73$ & $8.06$ \\
3 & $0.4236$ & $3.39$ & $15.60$ & $4.60$ & $9.19$ \\
4 & $0.4866$ & $3.89$ & $17.70$ & $4.55$ & $9.60$ \\
\bottomrule
\end{tabular}
\caption{Certified in-region bound at eight levels, mean absolute units per pixel and channel, five selected
checkpoints. The bound is below the edit magnitude for the two seeds with the smallest bound and above
it for the other three.}
\end{table}

\section{Post-hoc multiplicity sensitivity}

The within-HCOCO fMSE residuals are post-hoc sensitivity results, not evidence for the main
claim. H1 is the 12 within-HCOCO method-by-metric tests.
H5 is the broader family of all 48
estimable stratum records. The frozen four-method family excludes DiffHarmony, which is shown
only descriptively. The two families use the same raw bootstrap p values and different Holm
adjustments. Harmonizer and PCT-Net survive H1, while neither survives H5.
Table~S23 uses 20{,}000 bootstrap draws and Table~S18 uses 2{,}000, so the two tables give the
same point estimates with interval endpoints about one unit apart (Harmonizer fMSE
$-50.33$: $[-83.3, -17.4]$ here, $[-82.5, -16.2]$ in Table~S18). No sign changes.

\begin{table}[H]
\centering
\scriptsize
\begin{tabular}{lrrrrrrr}
\toprule
Method & Step & 95\% CI & Raw $p$ & H1 Holm & H1 & H5 Holm & H5 \\
\midrule
DoveNet & $-40.50$ & $[-122.0, +41.2]$ & $0.3311$ & $1.0000$ & no & $1.0000$ & no \\
Harmonizer & $-50.33$ & $[-83.3, -17.4]$ & $0.0028$ & $0.0252$ & yes & $0.1008$ & no \\
Editor (ours) & $-43.78$ & $[-99.0, +11.3]$ & $0.1242$ & $0.8694$ & no & $1.0000$ & no \\
PCT-Net & $-44.53$ & $[-73.6, -15.2]$ & $0.0024$ & $0.0240$ & yes & $0.0912$ & no \\
DucoNet & $-35.65$ & $[-66.9, -4.7]$ & $0.0240$ & $0.1920$ & no & $0.7920$ & no \\
DiffHarmony (descriptive) & $-78.15$ & $[-121.8, -35.8]$ & $0.0006$ & $0.0066$ & yes & $0.0234$ & yes \\
\bottomrule
\end{tabular}
\caption{Within-HCOCO fMSE exposure steps. H1 is the 12-row within-HCOCO family. H5 is the
48-row all-estimable-strata family. DiffHarmony is outside the frozen four-method family and is
descriptive only. Values were regenerated by the committed statistics-hardening script.}
\end{table}
\endgroup

\section{Harmonizer reconciliation}

\noindent The main text reports Harmonizer under one protocol throughout, the protocol of main-text Section~4: every composite, mask and target resized to $256 \times 256$, arguments
predicted on that resized composite. Its own paper reports the benchmark at the same resolution
(arXiv 2207.01322, Table 1), but its evaluation code predicts the filter arguments from the
native-resolution composite and resizes only afterwards. Running its protocol, with its released model
code and its own metric function, on the staged composites gives the published row.

\begin{table}[H]
\centering
\small
\begin{tabular}{l r r r}
\toprule
Quantity & Their protocol, our run & Published & Difference \\
\midrule
MSE (official list) & $23.99$ & $24.26$ & $1.10$ percent \\
fMSE (official list) & $277.02$ & $280.51$ & $1.24$ percent \\
PSNR (official list) & $37.88$ & $37.84$ & $0.04$ dB \\
MSE, our protocol & $107.08$ & n/a & their number under our setting \\
\bottomrule
\end{tabular}
\caption{The Harmonizer checkpoint is the one its paper describes. The $107.08$ quoted in the main
text's tables is that same checkpoint under our protocol, and the difference is the resolution at
which the arguments are predicted, not a different model.}
\label{tab:harmonizer_reconciliation}
\end{table}

\noindent The same choice moves the population contrast for this checkpoint. Under our protocol the
retained-to-excluded ratio is $5.05$ on MSE and $1.83$ on fMSE. Under its own it is $1.99$ and $0.99$,
so the apparent inflation is smaller under its authors' setting and absent on the foreground-region
metric. The population tables of the main text stay in one protocol, and this is the stated
sensitivity to it. Runs: three populations, eight CPU threads, about three hours wall clock, dominated
by the high-resolution sub-datasets because the argument prediction runs at native resolution. Records:
the summary files of the reconciliation under the release's notes, each with per-subset and pooled
values, per-sample rows and the published row it is compared against.

\section{Metric validation}

The dataset README reports two DoveNet rows: the paper model at MSE $52.36$, PSNR $34.75$\,dB and
fMSE $549.96$, and the released checkpoint we score at an expected MSE $50.87$ and PSNR
$34.83$\,dB, with no fMSE. Our pinned official-list numbers score the released checkpoint, so the
$52.36$ row is not the reference for them. Our PSNR averages per-image PSNR at a fixed peak of $255$, as
the published tables do, while the upstream scripts use each image's own output range as the peak,
which reads lower on the same images: the same list under that convention gives MSE $124.50$ and PSNR
$32.04$\,dB.

\emph{Data path.} Scoring the unharmonized composite with our loader and metrics over all 7{,}404
official composites gives MSE $170.06$, fMSE $1341.41$ and PSNR $31.96$\,dB against the README's
$172.47$, $1387.30$ and $31.63$\,dB, with per-subset MSE within $2.6$ percent and per-subset PSNR within $0.6$\,dB (largest deviations: HCOCO MSE $-2.5$ percent and Hday2night PSNR $+0.56$\,dB).
This fixes the test list, file pairing, resizing, colour handling, region of integration and pixel
scale as matching the published protocol.

\emph{BatchNorm mode.} The upstream test script, run as its README documents, never enables
evaluation mode and forces batch size one, so its batch normalization uses per-image statistics.
Our driver uses inference-mode statistics. Table~S25 reruns the released checkpoint over the full
official list under that upstream protocol (uint8 truncation, PSNR peak equal to each image's
output range).

\begin{table}[H]
\centering
\footnotesize
\begin{tabular}{lrrrrrrr}
\toprule
Subset & $n$ & our MSE & README & ratio & our PSNR & README PSNR & diff \\
\midrule
HCOCO & 4{,}283 & $33.61$ & $33.65$ & $0.999$ & $35.98$ & $35.95$ & $+0.03$ \\
HAdobe5k & 2{,}160 & $56.40$ & $55.99$ & $1.007$ & $34.23$ & $34.26$ & $-0.03$ \\
HFlickr & 828 & $125.92$ & $125.73$ & $1.002$ & $30.58$ & $30.58$ & $+0.00$ \\
Hday2night & 133 & $56.97$ & $56.26$ & $1.013$ & $34.73$ & $34.76$ & $-0.03$ \\
\midrule
overall & 7{,}404 & $51.00$ & $50.87$ & $1.003$ & $34.84$ & $34.83$ & $+0.01$ \\
\bottomrule
\end{tabular}
\caption{Released-checkpoint reproduction under the upstream protocol. The README column is the released
model's documented expectation, and the whole gap between our printed numbers and the published row is
the batch-normalization mode.}
\end{table}

\emph{Population contrast under both conventions.} The contrast the paper reports is present in both
modes and larger under the upstream one. Table~S26 gives the pooled populations and the HCOCO
restriction in inference mode (columns ``eval'') and training mode (columns ``train'').

\begin{table}[H]
\centering
\footnotesize
\begin{tabular}{lrrrrr}
\toprule
Pooled population & $n$ & MSE eval & MSE train & fMSE eval & fMSE train \\
\midrule
official (all) & 7{,}404 & $122.75$ & $49.99$ & $1132.68$ & $537.53$ \\
retained & 5{,}581 & $140.18$ & $58.21$ & $1114.71$ & $538.88$ \\
excluded & 1{,}823 & $69.40$ & $24.83$ & $1187.67$ & $533.39$ \\
ratio retained / excluded & & $2.02$ & $2.34$ & $0.94$ & $1.01$ \\
\bottomrule
\end{tabular}

\vspace{0.7em}
\begin{tabular}{lrrrrr}
\toprule
Inside HCOCO & $n$ & MSE eval & fMSE eval & MSE train & fMSE train \\
\midrule
retained & 2{,}476 & $98.96$ & $1240.67$ & $38.75$ & $554.04$ \\
excluded & 1{,}807 & $67.16$ & $1181.79$ & $24.46$ & $532.71$ \\
ratio retained / excluded & & $1.47$ & $1.05$ & $1.58$ & $1.04$ \\
\bottomrule
\end{tabular}
\caption{Population contrast under inference-mode (``eval'') and training-mode (``train'') batch
normalization, from the full-list run whose replication control passed (max $\lvert \Delta \mathrm{MSE}
\rvert$ $5.4 \times 10^{-5}$ against the stored records). The official list is a weighted average
over two differently composed populations: the excluded list is 99 percent HCOCO, the retained 44
percent. Nothing here licenses a causal exposure claim.}
\end{table}

Only the population-level ratios of Table~S26 are shown to hold under both conventions; the
foreground-ratio-matched step contrasts of main Table~2 were not recomputed in training mode.